\documentclass{article}
\usepackage{newtxtext}

\usepackage[main,final]{neurips_2026}

\usepackage[utf8]{inputenc} 
\usepackage[T1]{fontenc}    
\usepackage{hyperref}       
\usepackage{url}            
\usepackage{booktabs}       
\usepackage{amsfonts, amsmath, amsthm}       
\usepackage{nicefrac}       
\usepackage{microtype}      
\usepackage{xcolor}         
\usepackage{subcaption}
\newtheorem{theorem}{Theorem}
\newtheorem{lemma}{Lemma}

\newtheorem{definition}{Definition}

\newtheorem{assumption}{Assumption}
\newtheorem{remark}{Remark}
\usepackage{cleveref}
\usepackage{mathtools}

\newcommand{\R}{\mathbb{R}}
\newcommand{\E}{\mathbb{E}}
\newcommand{\That}{\hat{\theta}}

\newcommand{\Shat}{\hat{\Sigma}}

\newcommand{\norm}[1]{\left\lVert#1\right\rVert}

\DeclareMathOperator*{\argmin}{arg\,min}
\DeclareMathOperator{\diag}{diag}
\title{Reduced Cost Influence Functions for Predict-then-Optimize under Noisy Data}

\author{%
 William Zhang \\
  Operations Research Center\\
  Massachusetts Institute of Technology\\
  Cambridge, MA 02139 \\
  \texttt{wzhang64@mit.edu} \\
  \And
  Saurabh Amin \\
  Operations Research Center\\
  Massachusetts Institute of Technology\\
  Cambridge, MA 02139 \\
  \texttt{amins@mit.edu}\\
  \AND
  Georgia Perakis \\
  Operations Research Center\\
  Massachusetts Institute of Technology\\
  Cambridge, MA 02139 \\
  \texttt{georgiap@mit.edu} \\
}

\begin{document}

\maketitle

\begin{abstract}
  Machine learning models deployed in decision-making pipelines, from vehicle routing to dynamic pricing to clinical decision support, must remain reliable in the face of low-quality training data. We study a predict-then-optimize setting in which an upstream model predicts the cost vector of a downstream linear program, with a large pool of noisy training data supplemented by a small curated set of clean data. We introduce Reduced Cost Influence Functions (RCIFs), which combine influence functions from robust statistics with the geometry of linear optimization to estimate how each training point shifts downstream decisions. Building on this, we propose a reweighting algorithm that uses RCIFs to downweight harmful training points and reduce downstream regret. Experiments on random linear programs and shortest-path problems show that our approach reduces regret, and accompanying theory characterizes properties of RCIFs and identifies regimes favorable to our method relative to baselines.
\end{abstract}

\section{Introduction}
Machine learning models are increasingly used as components of larger decision-making pipelines: pricing engines, routing systems, clinical decision support tools. In these pipelines, predictions are not the end of the line: they are passed to a downstream optimization procedure whose solution carries the real consequences. This shift makes the training of predictive models a more delicate problem, since small errors under standard prediction loss can produce suboptimal downstream decisions. Compounding this, machine learning models are well known to be vulnerable to low-quality training data
\citep{biggio2018wild}, and a decision pipeline that is brittle to training-time corruption can produce costly errors downstream. Thus, how can we improve robustness of the downstream decision against corrupted training data via the upstream training process?

We propose a downstream-aware reweighting of the training data that
leaves both the predictor architecture and the downstream optimization method unchanged.
Our method assumes access to a small clean validation set, a standard assumption
in robust learning literature \citep{ren2018learning, hendrycks2018using}, in addition to a larger, potentially corrupted training set. We introduce the core technical object, the \emph{reduced cost influence function} (RCIF),
a decision-aware signal that propagates the effect of
upweighting each training point through the upstream predictor onto the reduced costs of the downstream LP, directly connecting each training point to the outcome decision. The RCIF measures how each training point destabilizes or stabilizes the downstream decision boundary; we use it to reweight the training data using the validation set, so that the refit predictor produces decisions that are robust to
training corruption.

Our method is best suited to predict-then-optimize problems in which
(i)~prediction errors frequently cause decision flips, (ii)~training
data is corrupted, and (iii)~a small clean validation set exists but is
too small to fit the predictor on its own. We demonstrate gains on random linear programs and on shortest-path routing, a standard benchmark for decision-making
pipelines \citep{elmachtoub2022smart}, and provide theoretical guarantees identifying validation-size regimes favorable to RCIF reweighting relative to both the standard, unweighted estimator and
validation-only estimator.

\section{Related Work}

\paragraph{Decision-aware learning.}
A large body of work studies training predictive models for downstream tasks. End-to-end approaches train the upstream predictor with respect to downstream decision quality, most notably via surrogate losses such as SPO+ \citep{elmachtoub2022smart} and the broader decision-aware learning literature \citep{hu2022fast,NEURIPS2022_0904c7ed, qi2025integrated, donti2017task, bertsimas2020predictive}. Our approach is complementary: rather than modifying the training loss, we reweight the training data using a downstream-aware signal, leaving both the predictor and the LP unchanged. This is similar to \citet{lawless2022note}, which directly reweights data via regret but does not consider corrupted data.

\paragraph{Robustness in decision-aware learning.}
A parallel line of work addresses robustness in decision-aware learning. Work such as \citet{sun2023predict, yeh2024end, patel2024conformal} consider robustness via sharper uncertainty
sets using conformal prediction, while
\citet{schutte2023robust, im2025smart} propose surrogate losses for robust
decision-focused learning. Instead, we directly reweight the training data, which preserves the original pipeline and focuses on robustness to the training phase rather than at deployment via robust downstreams. 

\paragraph{Reweighting under noisy data.}
Mitigating label corruption given access to a
small clean set (and reweighting data) is prevalent in robust learning
\citep{hendrycks2018using, zhang2020geometry, liu2021probabilistic}. Other work considers corruption detection such as \citep{northcutt2021confident,diakonikolas2019sever,sheikholeslami2021provably}. Notably, \citet{ren2018learning}
derive weights from influence functions (popularized in \citep{koh2017understanding, broderick2020automatic}) for robustness. We adopt the influence-function machinery but
propagate effects onto reduced costs of the downstream LP rather than
onto prediction loss, which differentiates our approach from a direct adaptation of validation-loss influence.
\section{Setting}
\label{setting}
We consider a predict-then-optimize pipeline in which an upstream predictive model produces an estimate of an unknown cost vector, which is then passed to a downstream linear program whose solution is the final decision.

\paragraph{Downstream optimization.} The downstream component solves
a linear program of the form
\[
    x^*(c) = \argmin_{x \in \mathcal{X}} c^\top x,
\]
where $c \in \mathbb{R}^d$ is the cost vector and $\mathcal{X} = \{x \in \mathbb{R}^d : Ax = b, \, x \geq 0\}$ is a closed, bounded polyhedron with diameter $L$ defined by a constraint matrix $A \in \mathbb{R}^{a \times d}$ and a right-hand side $b \in \mathbb{R}^a$. We assume throughout that the LP is nondegenerate, so that each optimal vertex corresponds to a unique basis.
\paragraph{Upstream predictor and data.}
The upstream component is a regression model
$f_\theta : \mathcal{Z} \to \mathbb{R}^d$ mapping features $z \in \mathcal{Z} \subseteq \mathbb{R}^q$ to predicted cost vectors, parameterized by $\theta \in \Theta$ with true parameter $\theta^\star \in \Theta$. We have access to two datasets: training data $\mathcal{S} = \{(z_i, \tilde c_i)\}_{i=1}^n$ with possibly corrupted labels, and clean validation data
$\mathcal{V} = \{(z_v, c_v)\}_{v=1}^m$ with uncorrupted labels, where typically $m \ll n$.

The label model is
\[
    c_i \;=\; f_{\theta^\star}(z_i) + \varepsilon_i,
\]
with $\varepsilon_i$ zero-mean $\sigma_\varepsilon$-subgaussian and mutually independent. A subset $\mathcal{C} \subseteq \{1,\ldots,n\}$ of size $|\mathcal{C}| = \beta n$, $\beta \in (0, 1/2)$, indexes the corrupted training points:
\[
    \tilde c_i \;=\; c_i + \eta_i \quad (i \in \mathcal{C}),
    \qquad
    \tilde c_i \;=\; c_i \quad (i \notin \mathcal{C}),
\]
where $\eta_i$ is an additional perturbation. The validation set $\mathcal{V}$ contains the clean costs $c_j$ by assumption. Throughout the paper, $\tilde c_i$ denotes the \emph{observed} training label at index $i$, which equals the clean label $c_i$ when $i \notin \mathcal{C}$. Equivalently, $\eta_i = 0$ for $i \notin \mathcal{C}$, so the decomposition $\tilde c_i = c_i + \eta_i$ holds for all $i$.
\paragraph{Fitting algorithm.} Given a loss function $\ell : \mathbb{R}^d \times \mathbb{R}^d \to \mathbb{R}$, the standard empirical risk minimizer (ERM) on the training data is
\[
    \hat\theta = \argmin_{\theta \in \Theta} 
    \sum_{i=1}^n \ell(f_\theta(z_i), \tilde c_i).
\]
For a given set of weights $p_1,\dots, p_n$, one can instead obtain the weighted ERM
\[
    \hat\theta_p = \argmin_{\theta \in \Theta} 
    \sum_{i=1}^n p_i \, \ell(f_\theta(z_i), \tilde c_i),
\]
with larger weights placing greater emphasis on the $i$-th training point.
We assume throughout that the empirical risk is strictly convex in $\theta$ and twice differentiable \citep{koh2017understanding}, so that influence-function approximations are well defined.

\section{Preliminaries}
\paragraph{Bases, reduced costs, and optimality cones.} A \emph{basis} $B \subseteq \{1, \ldots, d\}$ is a set of indices such that the submatrix $A_B$ is invertible and $A_B^{-1}b \geq 0$ is feasible, and $N(B) = \{1, \ldots, d\}
\setminus B$ denotes the corresponding nonbasic set. For each nonbasic index $k \in N(B)$, the \emph{edge direction} is a $d$-dimensional vector
\[
    \delta(B, k) =
    \begin{bmatrix} -A_B^{-1} A_k \\ e_k \end{bmatrix},
\]
where $e_k$ is the elementary unit vector in nonbasic coordinates corresponding to nonbasic index $k$.
The edge direction represents the direction along which the nonbasic variable $x_k$ can increase from zero while maintaining feasibility \citep{bennouna2026data,ye2026learning}. The \emph{reduced cost} of nonbasic variable $k$ under cost vector $c$ and basis $B$ is
\[
    s_k(c, B) = c^\top \delta(B, k),
\]
and $B$ is optimal for $c$ if and only if $s_k(c, B) \geq 0$ for all $k \in N(B)$. The set of cost vectors for which $B$ is optimal forms the \emph{optimality cone}
\[
    \Lambda(B) = \{c \in \mathbb{R}^d : s_k(c, B) \geq 0 \text{ for
    all } k \in N(B)\},
\]
and the cones $\{\Lambda(B)\}_B$ cover $\mathbb{R}^d$ with overlaps only on measure-zero boundaries. Two cost vectors in the same cone induce the same optimal decision.

\paragraph{Regret.} The quality of a predicted cost vector $\hat c$ relative to the true cost $c$ is measured by the downstream \emph{regret}:
\[
    \mathrm{Regret}(\hat c, c) = c^\top x^*(\hat c) - c^\top x^*(c),
\]
the excess true cost incurred by acting optimally under $\hat c$ rather than $c$. Regret is zero whenever $\hat c$ and $c$ lie in the same optimality cone, regardless of error $\|\hat c - c\|$.

\paragraph{Influence functions.} Influence functions \citep{koh2017understanding} provide a first-order
approximation to the effect of perturbing the training distribution on
the learned parameters. Consider the estimator obtained by upweighting
training point $w_j = (z_j, \tilde c_j)$ by $\epsilon$:
\[
    \hat\theta_{\epsilon, w_j} = \argmin_{\theta}
    \frac{1}{n} \sum_{i=1}^n \ell(f_\theta(z_i), \tilde c_i)
    + \epsilon \, \ell(f_\theta(z_j), \tilde c_j).
\]
Under our risk assumptions, the derivative of
$\hat\theta_{\epsilon, w_j}$ with respect to $\epsilon$ at $\epsilon =
0$ admits the closed form
\[
    \mathcal{I}_{\hat\theta}(w_j)
    := \left. \frac{d \hat\theta_{\epsilon, w_j}}{d \epsilon}
    \right|_{\epsilon = 0}
    = -H_{\hat\theta}^{-1} \nabla_\theta \ell(f_{\hat\theta}(z_j),
    \tilde c_j),
\]
where $H_{\hat\theta} = \frac{1}{n} \sum_{i=1}^n \nabla_\theta^2 \ell(f_{\hat\theta}(z_i), \tilde c_i)$ is the empirical Hessian of the training loss at $\hat\theta$. We refer to $\mathcal{I}_{\hat\theta} (w_j)$ as the \emph{influence} of training point $w_j$ on $\hat\theta$.

\section{Reduced cost influence function}
Recall that downstream regret is zero whenever the predicted cost $\hat c$ and the true cost $c$ lie in the same optimality cone. Thus the prediction need not be accurate in norm; it need only lie in the correct cone. Conversely, predictions near the boundary of $\Lambda(B)$ are \emph{unstable}: small perturbations of
$\hat c$ can push the prediction into an adjacent cone and induce a discontinuous jump in downstream cost.

The reduced cost $s_k(c, B)$ measures the slack in the $k$-th 
constraint defining $\Lambda(B)$. Geometrically, the distance from 
$ c$ to the $k$-th cone boundary equals $s_k(c, B)/\|\delta(B,k)\|$, 
and the overall margin is given by the minimum such distance over 
$k \in N(B)$ (this margin coincides with distance to degeneracy \citep{el2019generalization, liu2023active} under nondegeneracy). Our goal is to quantify the influence of each training point on this margin, identify harmful points (particularly in $\mathcal{C}$), and downweight them. To this end we introduce reduced cost influence functions (RCIFs).

Fix a validation input $z_v \in \mathcal{V}$ and let $\hat c_v =
f_{\hat\theta}(z_v)$ be the predicted cost at $z_v$. Let $B$ be the
optimal basis for $\hat c_v$, and let $k \in N(B)$ be a nonbasic index
of interest. We wish to quantify how upweighting training point $w_j$
changes the reduced cost $s_k(\hat c_v, B)$ at the validation point $z_v$.

Viewing $s_k$ as a function of $\theta$ through the predictor, we have
$s_k(f_\theta(z_v), B) = \delta(B, k)^\top f_\theta(z_v)$. Applying the
chain rule and evaluating at $\theta = \hat\theta$ in the influence
direction $\mathcal{I}_{\hat\theta}(w_j)$ yields the following
definition.

\begin{definition}[Reduced Cost Influence Function]
Let $\hat\theta$ be the ERM hypothesis, let $z_v$ be a validation input,
let $\hat c_v = f_{\hat\theta}(z_v)$, and let $B$ be the optimal basis for $\hat c_v$. For a nonbasic index $k \in N(B)$ and a training point $w_j$, the \emph{reduced cost influence function} is
\[
    \mathrm{RCIF}_{k, \hat\theta}(z_v, w_j)
    := \delta(B, k)^\top \, \nabla_\theta f_{\hat\theta}(z_v) \,
    \mathcal{I}_{\hat\theta}(w_j).
\]
\end{definition}

The RCIF admits a direct interpretation: it is the first-order
approximation of the change in the reduced cost $s_k(\hat c_v,
B)$ at validation point $z_v$ when training point $w_j$ is upweighted by
a unit of $\epsilon$. A positive RCIF means upweighting $w_j$
increases the reduced cost in direction $k$, pushing $\hat c_v$ deeper into
$\Lambda(B)$ along that facet. This signed, direction-aware signal is the basis for the
reweighting schemes introduced in the next section.

\subsection{RCIF score}
We use the RCIF to construct aggregate scores for reweighting and refitting the predictor. We present two score variants: a natural two-sided score and a second, more refined directed score.
\subsubsection{Two-sided weighting scheme}
For each validation point $w_v \in \mathcal{V}$, let $\hat c_v =
f_{\hat\theta}(z_v)$ be the predicted cost, $B_v$ the optimal basis
for $\hat c_v$, and define the most fragile nonbasic direction as
\[
    k^*_v = \argmin_{k \in N(B_v)} \frac{s_k(\hat c_v, B_v)}{\norm{\delta(B_v,k)}},
\]
i.e., the nonbasic index with the smallest reduced cost margin under
$\hat c_v$. This is the direction along which $\hat c_v$ is closest to
the boundary of $\Lambda(B_v)$, and thus the direction most
vulnerable to a basis flip.

We say validation point $w_v$ is \emph{correct} if $B_v$ is also the
optimal basis under the true cost $c_v$, i.e., if $\hat c_v$ and $c_v$
lie in the same optimality cone, and \emph{incorrect} otherwise. Let
$\sigma_v = 2 \cdot \mathbf{1}_{\text{correct}}(w_v) - 1 \in \{-1,
+1\}$ encode this status.

The two-sided aggregate score for training point $w_j$ is
\[
    \Psi_j^{\text{TS}}
    = \frac{1}{m}
    \sum_{w_v \in \mathcal{V}} \sigma_v \cdot
    \mathrm{RCIF}_{k^*_v, \hat\theta}(z_v, w_j).
\]

The sign $\sigma_v$ aligns two cases: at correct validation points we want to push $\hat c_v$ deeper into $\Lambda(B_v)$, and at incorrect points, we wish to destabilize the current basis so that $\hat c_v$ may exit the wrong cone.
\subsubsection{Directed weighting scheme}
\label{sec:reweighting}
The two-sided score destabilizes incorrect validation points along
the direction of smallest current margin, but this direction is not
necessarily the one that would move $\hat c_v$ toward the correct
cone. We have access to the
true cost $c_v$ and can identify the directions along which the true $c_v$ indicates a basis change is warranted: specifically, those
nonbasic indices $k \in N(B_v)$ for which $s_k(c_v, B_v) < 0$. These
are the directions in which the true cost violates the optimality
condition of the current basis $B_v$, and hence the directions whose
nonbasic variables should enter the basis under $c_v$.

We refine the choice of $k^*_v$ accordingly. For correct validation
points, we keep the most fragile direction as before. For incorrect
validation points, we restrict the argmin to the set of
true-cost-violating directions:
\[
    k^*_v =
    \begin{cases}
        \displaystyle\argmin_{k \in N(B_v)} \frac{s_k(\hat c_v, B_v)}{\norm{\delta(B_v,k)}},
        & \text{if } v \text{ is correct,} \\[1.5ex]
        \displaystyle\argmin_{k \in N(B_v) \,:\, s_k(c_v, B_v) < 0}
        \frac{s_k(\hat c_v, B_v)}{\norm{\delta(B_v,k)}},
        & \text{if } v \text{ is incorrect.}
    \end{cases}
\]
The directed aggregate score uses the same outer formula as the
two-sided score but with this refined choice of $k^*_v$:
\[
    \Psi^{\text{D}}_j
    = \frac{1}{m}
    \sum_{w_v \in \mathcal{V}} \sigma_v \cdot
    \mathrm{RCIF}_{k^*_v, \hat\theta}(z_v, w_j).
\]
The sign convention is unchanged. The refinement only changes \emph{which} direction is
destabilized. These scores can be used as additive weight adjustments directly, but an exponential refinement can improve performance in practice.
\subsection{Exponential reweighting}
Given an aggregate score $\Psi_j$ (either $\Psi_j^{\text{TS}}$
or $\Psi_j^{\text{D}}$) for each training point, we convert scores
to training weights via an exponential family with temperature
parameter $\lambda > 0$:
\[
    p_j = \frac{\exp(\lambda \Psi_j)}
               {\sum_{i=1}^n \exp(\lambda \Psi_{i})}.
\]
We then refit the predictor by solving the weighted ERM problem with weights $\{p_j\}$. 
The temperature hyperparameter $\lambda$ controls the aggressiveness of the
reweighting: $\lambda = 0$ recovers uniform weighting, while large
$\lambda$ concentrates mass on the highest-scoring training points. For experiments, we set $\lambda = 0.05$.

\subsection{Iterative reweighting}
A single round of RCIF reweighting evaluates the influence of each
training point at the original ERM estimator $\hat\theta$. After the
refit, however, the predictor has changed, and so have the optimal
bases at validation points and the most fragile (or directed) nonbasic
directions. We can sharpen the signal by recomputing the scores at
the refitted estimator and accumulating them across iterations.

For $\tau$ iterations, let $\Psi_j^{(t)}$ denote the aggregate score
of training point $w_j$ at the $t$-th iteration, computed using
$\hat\theta^{(t)}$, the predictor fitted at the start of iteration
$t$ (with $\hat\theta^{(1)}$ as unweighted ERM estimator). The
iterative weight for $w_j$ at iteration $\tau$ is
\[
    p_j^{(\tau)}
    = \frac{\exp\left(\lambda \sum_{t=1}^\tau \Psi_j^{(t)}\right)}
    {\sum_{i=1}^n \exp\left(\lambda \sum_{t=1}^\tau
    \Psi_{i}^{(t)}\right)}.
\]
Each iteration refits the predictor with weights $p_j^{(\tau)}$. We treat $\tau$
as a hyperparameter; in our experiments we set $\tau = 5$.
\section{Theoretical results}
We derive two properties of RCIF reweighting after one iteration of exponential weights: a signed score decrease for corrupted points, and validation-size regimes in which RCIF reweighting admits tighter regret upper bounds than the unweighted ERM and validation-only ERM baselines.
\subsection{Setting and assumptions}
We consider the same predict-then-optimize setting as defined in \Cref{setting}, but with a linear upstream model such that for parameter $\theta \in \mathbb{R}^{d\times q}$, $z\in \mathcal{R}^q$, $f_{\theta}(z) = \theta z$, and the loss function is square loss $\ell(c,\hat c) = \|c - \hat c\|^2$. Let $\hat\theta$ denote standard ERM on training set $\mathcal{S}$, $\hat\theta_p$ denote the one-step RCIF-reweighted ERM, and $\hat\theta_{\mathrm{val}}$ denote the validation-only ERM fit on $\mathcal V$.
Unless stated otherwise, throughout this section, we make the following assumptions.

\begin{description}
    \item[(D1)] \textbf{Bounded Features.} Training and validation points $(z,c)$ are drawn iid from distribution $\mathcal{D}$, with $\|z\|\leq R$. Also let empirical design covariance $\hat\Sigma := \frac{1}{n}\sum_{i=1}^{n} z_i z_i^\top \in \R^{q \times q}$ be invertible and 
satisfy $\lambda_{\min}^{\mathrm{eig}}(\hat\Sigma) \geq \mu > 0$.
    \item[(D2)] \textbf{Labels and noise.} For both training and validation, for any $i$, for true parameter $\theta^*$, $c_i = \theta^\star z_i + \varepsilon_i$ with each entry of $\varepsilon_i$
    0-mean $\sigma_{\varepsilon}$-sub-Gaussian, independent of each other and features.

    \item[(D3)] \textbf{Corruption set.} A set $\mathcal{C} \subset
    \{1,\ldots,n\}$ of size $|\mathcal{C}| = \beta n$ is drawn
    uniformly at random, independently of features and noise, with
    $\beta \in (0, 1/2)$.

    \item[(D4)] \textbf{Bounded directional corruption.} For $j \in \mathcal{C}$, the corruption is of the form
    $\tilde c_j = c_j + \eta_j$ where $\eta_j = - \alpha\,\hat\delta_j$, and where $\hat\delta_j :=
    \delta(B_j^\star, k_j)/\|\delta(B_j^\star, k_j)\|$ is
    the unit-norm direction at the optimal basis $B_j^\star$ at
    $c_j$ for non-basic direction $k_j \in N(B_j^\star)$ chosen independently of validation sample or learned estimator.

    \item[(D5)] \textbf{Bounded LP geometry.} $\|\delta(B, k)\| \leq D$
    for every basis $B$ and nonbasic $k$.
\end{description}
\subsection{RCIF score decomposition}
We present a factorization of the RCIF, which leads to a score decomposition that isolates a ``clean" RCIF score and a ``corrupted" RCIF score from corruption $\eta_j$.
\begin{lemma}[RCIF Score Factorization]
\label{lem:factorization}
    We can factor the RCIF of training point $w_j$ on a single validation input $z_v$ on non-basic direction $k$ into two different terms, a downstream residual interaction term, and an input alignment term. Then we have that 
    \begin{align*}
        \mathrm{RCIF}_{k, \hat\theta}(z_v, w_j) = -\underbrace{\delta(B_v, k)^\top r_j}_{\textsc{residual interaction}} \cdot \underbrace{\kappa(z_v,z_j)}_{\textsc{input alignment}}
    \end{align*}
    where $\kappa(z_v, z_j)$ denotes Mahalanobis inner product $z_v^{T} \Shat^{-1} z_j$ and $r_j$ denotes the residual of training point $j$, $r_j \coloneqq \That z_j - \tilde{c}_j$.
\end{lemma}
Now we may show that the RCIF score can be decomposed into a clean and corrupted term.
\begin{lemma}[RCIF Score Decomposition]
\label{lem:score_decomp}
    For fixed $\hat \theta$ and training point $w_j = (z_j, \tilde c_j) = (z_j,c_j + \eta_j)$, we can decompose the RCIF score $\Psi_j$ into a clean score and noise term:
    $\Psi_j = \Psi_j^{\mathrm{clean}} + u_j^\top \eta_j$
    where $u_j = \frac{1}{m}\sum_{w_v \in V}
    \sigma_v\, \kappa(z_j, z_v) \delta(B_v, k^*_v)$, where $k^*$ is the appropriate direction depending on the weighting scheme, and $\Psi_j^{\mathrm{clean}}$ is the score if the score were calculated using $w_j = (z_j,c_j)$.
\end{lemma}
These lemmas hold without Assumptions (D1)-(D5) (handling any $\eta_j$, not just the one in D4).
Via this decomposition, we can obtain results that show how scoring on corrupted points differs between clean and corrupted points, with differences in magnitude under zero-mean corruption (see Appendix \ref{sec:scoremag}) as well as signed score decrease for directional corruption as in D4.
\subsection{Decreased corruption score}
We show that under certain settings, our scoring method will give a lower score to corrupted points. First we define an alignment parameter.
\begin{definition}[Validation alignment]\label{def:align}
    For a fresh sample $(z_j,c_j)$, and fixed $\That$, conditional on $(\That, z_j, c_j)$, define the \emph{alignment parameter}
    \[
        \gamma(\That, z_j, c_j) \;\coloneqq\; \E_{w_v \sim \mathcal{D}}\!\left[\sigma_v\kappa(z_j, z_v)\delta(B_v, k^*_v)^\top \hat\delta_j \,\middle|\, \That, z_j, c_j\right].
    \]
\end{definition}
\begin{theorem}[Signed score decrease under alignment; informal]\label{thm:signed-downweight-informal}
Adopt Assumptions (D1)--(D5). Let $(z_j, c_j)$ be a fresh sample drawn from $\mathcal{D}$
independently of $\hat\theta$ and $\mathcal{V}$, and let $\tilde c_j = c_j + \eta_j$ be its
directional corruption (D4). Suppose $\gamma := \gamma(\hat\theta, z_j, c_j) > 0$. Then,
in expectation over $\mathcal{V}$ conditional on $(\hat\theta, z_j, c_j)$, the score on the
corrupted point is shifted by $-\alpha\gamma$ relative to its clean counterpart, and with
probability at least $1 - \exp(-m\gamma^2/(8M_\gamma^2))$ over the same randomness, for constant $M_\gamma$,
\[
    \Psi_j(\tilde c_j) - \Psi_j(c_j) \;\leq\; -\frac{\alpha \gamma}{2}.
\]
The full statement, constants, and
proof appear as Theorem~\ref{thm:signed-downweight} in Appendix~\ref{app:downweight}.
\end{theorem}
\begin{remark}
    The assumption that $\gamma>0$ is a natural one. Note that $\sigma_v>0$ holds if $\That$ is similar to $\theta^*$, i.e. the upstream model makes predictions in the same cone. Furthermore, $\kappa(z_v,z_j)>0$ is likely positive if the distribution is tight around $\bar z = \mathbb{E}[z] \neq 0$, with large $n$. A tight distribution also makes $ \delta(B_v, k^*_v)^\top \hat\delta_j > 0$ likely, as it holds when a large proportion of the cost-vector validation distribution falls into a single optimality cone with the same fragile direction. 
\end{remark}
Thus, if a freshly sampled point is corrupted via Assumption D4, w.h.p. its RCIF score is decreased compared to its clean counterpart. This requires the point to be out-of-sample, but we show an in-sample counterpart with an additional $O(1/n)$ slack in Appendix~\ref{app:insample}. 
\subsection{Expected regret upper-bound comparison}
Under further assumptions, we can obtain regimes on validation set size $m$ in which the RCIF reweighted solution $\That_p$, has tighter upper-bounds on the expected regret than those of the standard ERM solution $\That$, and the validation-only solution $\That_{\mathrm{val}}$. For intuition, we include two relevant assumptions below.
\begin{assumption}[High probability alignment]
\label{ass:sb}
There exist constants $\gamma_0 > 0$ and $\delta_\gamma \in (0, 1/n)$
such that
\[
    \Pr\bigl[\gamma(\hat\theta, z_j, c_j) \geq \gamma_0\bigr] \;\geq\; 1 - \delta_\gamma,
\]
where the probability is over $(\hat\theta, z_j, c_j)$ with $\hat\theta$ fit on iid samples drawn from $\mathcal{D}$ (D1) and $(z_j,c_j)$ independently drawn from $\mathcal{D}$.
\end{assumption}
The assumption says the conditional alignment is bounded away from
zero with high probability.
To obtain a result on regret, we require the following condition. Furthermore, let $M$ be an RCIF score upper-bound $|\Psi_j| \leq M$, which holds under (D1)-(D5) with bounded residuals.
\begin{assumption}[Margin]
\label{def:margin}
There exists a constant $\omega > 0$ such that
\[
    \pi_v(\theta^\star z) \;\geq\; 1/\omega
    \quad\text{almost surely under } w=(z,c)\sim\mathcal{D},
\]
where $\pi_v(c) := \min_{k \in N(B^\star(c))}
s_k(c, B^\star(c))/\|\delta(B^\star(c), k)\|$ is the reduced-cost
margin at $c$.
\end{assumption}
\begin{theorem}[Validation-size regret regimes; informal]\label{thm:regret-regimes-informal}
Adopt Assumptions (D1)--(D5), Assumption~\ref{ass:sb}, and Assumption~\ref{def:margin}, and additional regularity assumptions listed in Appendix ~\ref{app:regret}.
There exist thresholds $m_\star(\lambda), m^\star(\lambda) \in [1, \infty]$, depending on
the problem constants (e.g., $\sigma_\varepsilon, \alpha, \beta, R, \mu, D, \gamma_0, \omega$), such that
in the asymptotic regime $n \to \infty$:
\begin{enumerate}
    \item (\emph{Lower threshold $m_\star$.}) For all $m > m_\star(\lambda)$, the RCIF
    reweighted estimator $\hat\theta_p$ admits a tighter expected regret upper bound
    than the unweighted ERM $\hat\theta$. The threshold is finite whenever the
    contraction factor $\Phi(\lambda, m, n)$ from Lemma~\ref{lem:weight-ratio}
    eventually drops below $\beta \cdot A$ (for some constant $A$) in $m$.

    \item (\emph{Upper threshold $m^\star$.}) For all $m < m^\star(\lambda)$,
    $\hat\theta_p$ admits a tighter expected regret upper bound than the
    validation-only ERM $\hat\theta_{\mathrm{val}}$. The threshold is finite,
    since the irreducible corruption bias of $\hat\theta_p$ eventually exceeds
    val-only's $1/m$ variance.

    \item (\emph{Sweet spot.}) When $m_\star(\lambda) < m^\star(\lambda)$, every $m$
    in this interval admits a tighter expected regret upper bound for $\hat\theta_p$
    than for both baselines simultaneously.
\end{enumerate}
The thresholds and the sweet-spot condition are characterized explicitly in terms of
$\Phi(\lambda, m, n)$ in Theorem~\ref{thm:regret-regimes} (Appendix~\ref{app:regret}),
where the full statement, constants, and proof appear.
\end{theorem}

\begin{remark}
Theorem~\ref{thm:regret-regimes-informal} compares upper bounds on
expected regret, not the regrets themselves. However, the thresholds
indicate a transition region in validation size: small $m$ favors
training-data methods (unweighted ERM), large $m$ favors
validation-only ERM, and RCIF reduces the bias enough to compete
with both in the intermediate regime as observed in Section~\ref{sec:experiments}.
\end{remark}

\section{Experimental results}
\label{sec:experiments}

We evaluate our reweighting scheme across several settings, measuring
out-of-sample regret on a clean test set when the training data is
corrupted but a small clean validation set is available. To simulate
label corruption, we uniformly sample a fraction $\beta$ (default
$0.3$) of the training points and add isotropic Gaussian noise to
their cost labels; the validation and test sets remain clean.

\paragraph{Methods.} We compare ten methods.
\textbf{Iterative Exp.\ Directed} (ours): iterative
exponential reweighting with directed score $\Psi^{\text{D}}$.
\textbf{Exp.\ Two-Sided} (ours): single-round exponential
reweighting with $\Psi^{\text{TS}}$.
\textbf{Directed} (ours): additive reweighting
$p_i = 1/n + \phi\, \Psi^{\text{D}}_i$ with $\phi = 0.05$.
\textbf{Unweighted}: ERM on the corrupted training set.
\textbf{RC Weight}: single-round exponential reweighting using the raw
reduced-cost margin $\Psi^{\text{RC}}_j = \min_{k \in
N(B_j)} s_k(\hat{c}_j, B_j)$ as score.
\textbf{Validation Only}: ERM on the clean validation set alone.
\textbf{SPO Influence Weight}: reweighting via influence functions on the SPO+ surrogate loss rather than reduced costs; the closest analog to our method, isolating the
contribution of reduced costs over a surrogate loss (Appendix~\ref{app:spo-baseline}).
\textbf{SPO+ Validation Only}: an end-to-end predictor trained by subgradient descent on the SPO+ loss on the validation set alone, initialized at the validation-only least-squares fit.
\textbf{SPO+ Finetune}: unweighted ERM on the noisy train-plus-validation pool, fine-tuned by SPO+ subgradient descent on the validation set.
\textbf{Clean ERM (oracle)}: ERM on the full uncorrupted dataset, a reference lower bound on achievable regret.
For all methods except the validation-only ones, we fold the clean validation set into the second-stage fit, assigning each validation point the maximum weight
given to any training point, so every method using training data sees the same
total data. Additional experiments and details appear in Appendix~\ref{app:exp}.

\subsection{Random linear program}
We sample a random constraint matrix $A$ and right-hand side $b$ defining the downstream LP, and a random linear map $\theta^*$ from
features to costs. Training, validation, and test inputs are drawn i.i.d., with cost labels $c = \theta^* z + \varepsilon$. The upstream
predictor is a linear model fit by squared loss. Table~\ref{tab:results_new_val}
reports results with upstream dimension $20$, LP dimension $10$, $20$ constraints, $200$ training points, $50$ validation points,
$150$ test points, and corruption rate $\beta = 0.3$. Iterative Exp. Directed achieves the lowest regret, the gap to unweighted is modest in this setting but significant under a two-sided paired Wilcoxon signed-rank test over seeds ($p$ = 0.007).

\begin{table}[ht]
\centering
\small
\begin{tabular}{lccc}
\toprule
Method & Random LP & Shortest Path & NN Upstream \\
\midrule
Iterative Exp. Directed & \textbf{0.3451} $\pm$ \textbf{0.0133} & \textbf{0.9437} $\pm$ \textbf{0.0172} & 1.4511 $\pm$ 0.0187 \\
Exp. Two-Sided          & 0.3535 $\pm$ 0.0129 & 1.0009 $\pm$ 0.0175 & 1.4211 $\pm$ 0.0196 \\
Directed                & 0.3531 $\pm$ 0.0129 & 0.9825 $\pm$ 0.0171 & \textbf{1.3958} $\pm$ \textbf{0.0217} \\
Unweighted              & 0.3564 $\pm$ 0.0130 & 1.0029 $\pm$ 0.0169 & 1.4317 $\pm$ 0.0198 \\
RC Weight               & 0.3549 $\pm$ 0.0131 & 1.0035 $\pm$ 0.0169 & 1.4255 $\pm$ 0.0227 \\
Validation Only         & 0.4184 $\pm$ 0.0191 & 0.9989 $\pm$ 0.0181 & 1.4208 $\pm$ 0.0228 \\
SPO Influence Weight    & 0.3509 $\pm$ 0.0124 & 0.9560 $\pm$ 0.0158 & 1.4626 $\pm$ 0.0183 \\
Clean ERM (oracle) & 0.2785 $\pm$ 0.0109 & 0.8513 $\pm$ 0.0171 & 1.3452 $\pm$ 0.0190 \\
SPO+ Val Only & 0.4854 $\pm$ 0.0216 & 1.0914 $\pm$ 0.0182 & -- \\
SPO+ Finetune & 0.3666 $\pm$ 0.0140 & 1.0042 $\pm$ 0.0176 & -- \\
\bottomrule
\end{tabular}
\caption{Mean test regret over 100 seeds under
random LP setting, shortest path
setting with a linear upstream, and shortest path setting with a neural-network upstream. Best per column in bold.}
\label{tab:results_new_val}
\end{table}
\subsection{Shortest path setting}
For a more structured downstream LP, we consider the shortest-path problem on a grid network, following \citet{elmachtoub2022smart}: traverse from the northwest to the southeast corner of the grid. We use the PyEPO library \citep{tang2022pyepo} to generate the dataset with an upstream linear model fit via squared loss. Table~\ref{tab:results_new_val} reports results on a $4 \times 4$ grid with upstream input dimension $30$, $400$ training points, $100$ validation points, $200$ test points, and $\beta = 0.2$. On this more structured problem, RCIF reweighting yields substantially larger gains: Iterative Exp. Directed again performs best, by a much wider margin than in the random LP setting (paired $p$ $< 0.048$ for all methods). Furthermore, though the raw magnitude difference between methods is small, the recovery relative to Clean ERM is large: our method recovers 39\% of the error induced by noise (0.152 between Clean ERM and Unweighted) whereas the next best method SPO Influence Weight only recovers 30\%.
\subsection{Neural network upstream}
To demonstrate that the method retains value when the upstream predictor is nonlinear, we replace the linear model with a two-layer fully-connected network with hidden dimension $8$ and Tanh activation as seen in the last column of \Cref{tab:results_new_val}. We use the shortest path setting again, with a $5 \times 4$ grid, upstream input dimension $30$, $200$ training points, $100$ validation points, $200$ test points, and $\beta = 0.3$. We compare methods that leave the training procedure unchanged (reweighting and the clean-label reference); the SPO+ Finetune and SPO+ Validation Only baselines are defined for the linear upstream and require a specialized pipeline, so we omit them. Here the additive Directed variant performs best. We conjecture that nonlinearity makes the iterations of Iterative Exp.\ Directed harmful, since influence-function approximation error compounds across iterations, consistent with the known fragility of influence functions in nonlinear settings \citep{basu2020influence}. Correspondingly, the non-iterative Exp.\ Two-Sided performs relatively better here than in the linear settings, and the Directed variant yields strong improvements through more targeted scoring without added error.
\subsection{Validation size study}
Now we connect back to our theoretical result in \Cref{thm:regret-regimes-informal}, which provides conditions on the validation set size for which our method has an improved regret upper bound by providing a study on validation set size on shortest path problems (4$\times$4 grid, upstream input dimension 20, 500 training points, 200 test points, corruption rate $\beta=0.3$, 100 seeds with validation set varied over $\{5,10,20,50,100,150\}$) and observing that it translates to real regret improvements. As the size of the validation set increases, at first, the unweighted method $\hat \theta$ outperforms our method, but then for sufficient validation data, our method $\hat \theta_{\mathrm{p}}$ wins, but then finally at large validation, $\hat{\theta}_{\mathrm{val}}$ wins. We can visually see this sweet spot where our method wins in \Cref{fig:valsize} on moderate validation set sizes, demonstrating its value in these regimes. See Appendix~\ref{app:ablvalfull} for details.
\begin{figure}
    \centering
    \includegraphics[width=0.45\linewidth]{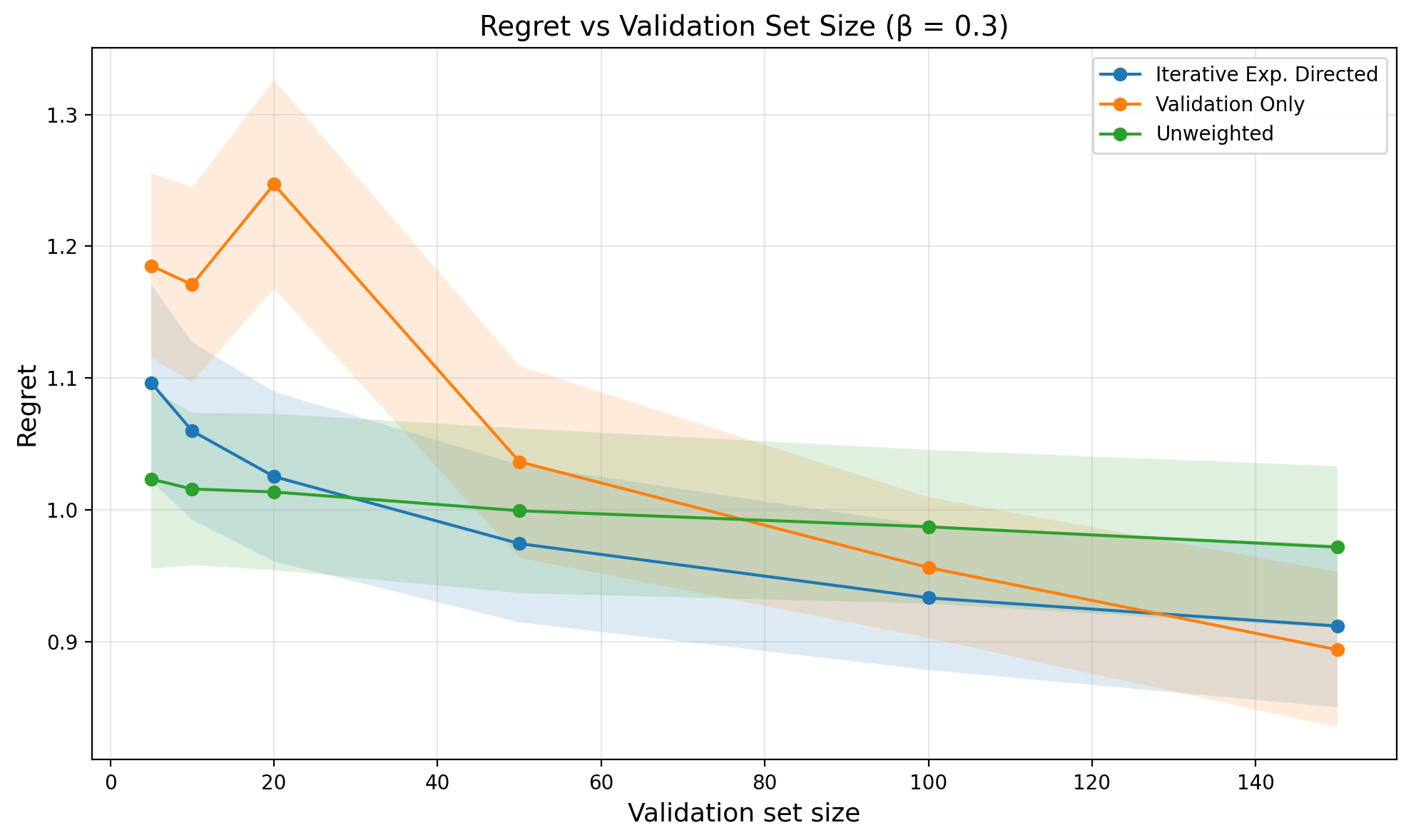}
    \caption{Test regret vs. validation size for shortest path on 4$\times$4 grid (setup in text).}
    \label{fig:valsize}
\end{figure}
\subsection{Misspecification}
Similarly to the neural network setting, we test what happens when the upstream model is misspecified. We keep the shortest-path setting ($4 \times 4$ grid, input dimension $30$, $400/100/200$ train/validation/test points, $\beta = 0.2$, $50$ seeds) and raise the polynomial degree of PyEPO's feature-to-cost map from $1$ to $2$ and $4$, so that the true costs are increasingly nonlinear in the features while the predictor remains linear and is fit by squared loss. We report results in \Cref{tab:misspec}. Raw regret values are not comparable across degrees, since PyEPO rescales the costs with the degree, so each column should be read on its own (degree 1 coincides with \Cref{tab:results_new_val}). The gains of RCIF reweighting are preserved and, relative to the corruption damage, grow proportionally with degree: Iterative Exp.\ Directed attains a test regret of $0.4856 \pm 0.0135$ at degree $2$ and $0.2899 \pm 0.0115$ at degree $4$, against $0.5715 \pm 0.0145$ and $0.3483 \pm 0.0135$ for unweighted ERM, and it has lower regret than every other competing method. RCIF scores only require the LP geometry at the predicted cost and the first-order response of the predictor, neither of which directly depends on the model being correctly specified, so the mechanism continues to identify harmful training points when the linear fit is only an approximation.
\begin{table}
    \centering
    \small
    \begin{tabular}{lccc}
        \toprule
        Method & deg = 1 & deg = 2 & deg = 4 \\
        \midrule
        Iterative Exp. Directed & \textbf{0.9437} $\pm$ \textbf{0.0172} & \textbf{0.4856} $\pm$ \textbf{0.0135} & \textbf{0.2899} $\pm$ \textbf{0.0115} \\
        Exp. Two-Sided & 1.0009 $\pm$ 0.0175 & 0.5682 $\pm$ 0.0152 & 0.3393 $\pm$ 0.0127 \\
        Directed & 0.9825 $\pm$ 0.0171 & 0.5347 $\pm$ 0.0132 & 0.3208 $\pm$ 0.0122 \\
        Unweighted & 1.0029 $\pm$ 0.0169 & 0.5715 $\pm$ 0.0145 & 0.3483 $\pm$ 0.0135 \\
        RC Weight & 1.0035 $\pm$ 0.0169 & 0.5713 $\pm$ 0.0145 & 0.3496 $\pm$ 0.0138 \\
        Validation Only & 0.9989 $\pm$ 0.0181 & 0.5284 $\pm$ 0.0146 & 0.3782 $\pm$ 0.0147 \\
        SPO Influence Weight & 0.9560 $\pm$ 0.0158 & 0.4999 $\pm$ 0.0119 & 0.2975 $\pm$ 0.0117 \\
        Clean ERM (oracle) & 0.8513 $\pm$ 0.0171 & 0.3972 $\pm$ 0.0118 & 0.2340 $\pm$ 0.0113 \\
        SPO+ Val Only & 1.0914 $\pm$ 0.0182 & 0.6857 $\pm$ 0.0165 & 0.4410 $\pm$ 0.0160 \\
        SPO+ Finetune & 1.0042 $\pm$ 0.0176 & 0.5406 $\pm$ 0.0151 & 0.3200 $\pm$ 0.0123 \\
        \bottomrule
        \end{tabular}
    \caption{Mean test regret, 50 seeds, shortest path across polynomial degree for feature-cost mapping.}
    \label{tab:misspec}
\end{table}
\section{Conclusion}
We introduced reduced cost influence functions, a downstream-aware signal that propagates training-point influence through the upstream predictor onto the reduced costs of the LP. Building on a score decomposition that isolates clean from corruption-induced contributions, we showed that RCIF scoring decreases the score of directionally corrupted points with high probability, and that there exist validation-size regimes in which RCIF reweighting admits tighter expected-regret upper bounds than unweighted and validation-only ERM. Experiments on random LPs, shortest-path routing, and a neural-network upstream are consistent with these regimes empirically and show consistent gains, particularly for shortest path. The neural-network setting also reveals a limitation: iterating exponential reweighting compounds the approximation error when the upstream is nonlinear, but the non-iterative variant still maintains strong performance. Furthermore, the downstream problem is currently limited to LPs, with assumed nondegeneracy for theoretical results. Promising directions include relaxing the nondegeneracy and connecting with the distance to degeneracy concept, extending the framework to MILPs and convex programs, and combining RCIF scoring with robust regression and active learning techniques for data selection.

\bibliographystyle{plainnat}
\bibliography{biblio}
\appendix

\newpage
\section{Technical appendices and supplementary material}
\subsection{Proof of \Cref{lem:factorization}}
\textbf{RCIF score factorization}
\begin{proof}
    By definition of the influence function for linear models, 
    the Hessian of the average training loss, viewed as an operator on
    $\R^{d \times q}$ (vectorized in row-major order), is
    \[
        H_{\That}
        = \frac{1}{n}\sum_{i=1}^n
          \nabla_\theta^2 \|\theta z_i - \tilde{c}_i\|^2
        = 2\,I_d \otimes \Shat,
    \]
    since the $(l_1,m_1;\,l_2,m_2)$ entry of the Hessian is
    $2\,\mathbf{1}[l_1=l_2]\,(z_i)_{m_1}(z_i)_{m_2}$, and summing gives
    $2\,\mathbf{1}[l_1=l_2]\,\Shat_{m_1 m_2}$.
    The inverse is $H_{\That}^{-1} = \tfrac{1}{2}\,I_d \otimes \Shat^{-1}$.
    Let $\operatorname{vec}$ denote row-major vectorization so $\operatorname{vec}(r_j z_j^\top) = r_j \otimes z_j$.
    The influence function is
    \begin{align*}
        \operatorname{vec}\bigl(I_{\That}(w_j)\bigr)
        &= -H_{\That}^{-1}\operatorname{vec}\bigl(\nabla_\theta\ell\bigr) \\
        &= -\tfrac{1}{2}(I_d \otimes \Shat^{-1})
            \cdot 2\bigl[r_j \otimes z_j\bigr] \\
        &= -(I_d\, r_j) \otimes (\Shat^{-1} z_j) \\
        &= -r_j \otimes (\Shat^{-1} z_j),
    \end{align*}
    where we used the Kronecker product rule
    $(A \otimes B)(u \otimes v) = Au \otimes Bv$.
    Reshaping back to $\R^{d \times q}$ gives
    \[
        I_{\That}(w_j) = -r_j\,(\Shat^{-1} z_j)^\top.
    \]
    Thus by definition of the RCIF, we have 
    \begin{align*}
    \mathrm{RCIF}_{k, \hat\theta}(z_v, w_j)
    &= \delta(B_v, k)^\top \, \nabla_\theta f_{\hat\theta}(z_v) \,
    \mathcal{I}_{\hat\theta}(w_j)\\&=\delta(B_v, k)^\top \, \,
    \mathcal{I}_{\hat\theta}(w_j) z_v\\&= -\delta(B_v, k)^\top \, r_j\,(\Shat^{-1} z_j)^\top z_v\\
     &= -\delta(B_v, k)^\top \, r_j \kappa(z_v, z_j)
    \end{align*}
    where the second equality follows by the directional derivative of a linear function, and the third equality follows by plugging in the derived influence function. 
\end{proof}
\subsection{Proof of \Cref{lem:score_decomp}}
\textbf{RCIF score decomposition}
\begin{proof}
    By \Cref{lem:factorization}, we can write the RCIF score on corrupted point $w_j$: 
    \[\Psi_j
    \;\coloneqq\;
    \frac{1}{m}\sum_{w_v \in V}
    \sigma_v\,
    \mathrm{RCIF}_{k^*_v,\,\That}\!\bigl(z_v, w_j\bigr)
    \;=\;
    -\frac{1}{m}\sum_{w_v \in V}
    \sigma_v\, \delta(B_v, k^*_v)^\top r_j \;
    \kappa(z_j, z_v). \]
    Note that $r_j = \hat{\theta}z_j - c_j - \eta_j = r_j^{\text{clean}} - \eta_j$ where $r_j^{\text{clean}}$ is the residual if the point was uncorrupted thus
    \begin{align*}
        -\frac{1}{m}\sum_{w_v \in V}
    \sigma_v\, \delta(B_v, k^*_v)^\top r_j \;
    \kappa(z_j, z_v) &= -\frac{1}{m}\sum_{w_v \in V}
    \sigma_v\, \delta(B_v, k^*_v)^\top (r_j^{\text{clean}} - \eta_j)  \;
    \kappa(z_j, z_v) \\ &=  -\frac{1}{m}\sum_{w_v \in V}
    \sigma_v\, \delta(B_v, k^*_v)^\top r_j^{\text{clean}}   \;
    \kappa(z_j, z_v) +  \frac{1}{m}\sum_{w_v \in V}
    \sigma_v\, \delta(B_v, k^*_v)^\top \eta_j  \;
    \kappa(z_j, z_v)\\ &= \Psi_j^{clean}  + \left(\frac{1}{m}\sum_{w_v \in V}
    \sigma_v\, \kappa(z_j, z_v) \delta(B_v, k^*_v)\right)^\top \eta_j    \\ &= \Psi_j^{clean}  + u_j^\top \eta_j
    \end{align*}
    so we see that the score on a corrupted point decomposes into an uncorrupted score and an additional noise term.
\end{proof}
\subsection{Score difference in magnitude for corrupted points}
\label{sec:scoremag}
Instead of the directed corruption paradigm in Assumption D4, we use the following:

For $j \in \mathcal{C}$, $\tilde c_j = c_j + \eta_j$ where $\eta_j$ is an additional bounded perturbation such that $\eta_j$ is zero-mean and independent from $\mathcal{S}, \mathcal{V}$ and across indices. 

We show that the RCIF score magnitude for a fresh sample (independent of $\mathcal{S}, \mathcal{V}$) that has been corrupted $(z_j,\tilde c_j) = (z_j,c_j + \eta_j)$ is larger than the magnitude for uncorrupted version on $(z_j, c_j)$. 
\begin{theorem}[Magnitude separation in expectation]\label{thm:mag-sep}
    Let $(z_j, c_j)$ be a fresh sample with $c_j = \theta^* z_j + \varepsilon_j$,
    independent of $\mathcal{S}$ used to fit $\hat\theta$. Assume the
    corruption $\eta_j$ satisfies
    \[
        \E[\eta_j \mid \mathcal{G}] = 0, \qquad
        \text{Cov}(\eta_j \mid \mathcal{G}) \succeq \sigma_\eta^2 \, I,
    \]
    where $\sigma$-algebra $\mathcal{G} = \sigma(\hat\theta,\mathcal{V},z_j,c_j)$. Let $\Psi^{\mathrm{clean}}_j$ be the score evaluated at the clean label $(z_j, c_j)$ and let $\Psi_j$ be the score evaluated at $(z_j, \tilde c_j) = (z_j, c_j + \eta_j)$. Then
    \[
        \E\!\left[\Psi_j^2 - (\Psi_j^{\mathrm{clean}})^2\right]
        \;\geq\; \sigma_\eta^2 \, \E\|u_j\|^2 \;\geq\; 0.
    \]
    where under most distributions the last inequality is strict positivity.
\end{theorem}
\begin{proof}
    By the score decomposition (Lemma~\ref{lem:score_decomp}),
    $\Psi_j = \Psi_j^{\mathrm{clean}} + u_j^\top \eta_j$. Let
    $\Delta_j = u_j^\top \eta_j$, so
    \[
        \Psi_j^2 - (\Psi_j^{\mathrm{clean}})^2
        \;=\; 2\,\Psi_j^{\mathrm{clean}}\,\Delta_j + \Delta_j^2.
    \]
    Since $\Psi_j^{\mathrm{clean}}$ and $u_j$ are $\mathcal{G}$-measurable
    and $\E[\eta_j\mid\mathcal{G}]=0$,
    \[
        \E[\Psi_j^{\mathrm{clean}}\,\Delta_j \mid \mathcal{G}]
        \;=\; \Psi_j^{\mathrm{clean}}\,u_j^\top \E[\eta_j\mid\mathcal{G}]
        \;=\; 0,
    \]
    and by the covariance hypothesis,
    \[
        \E[\Delta_j^2 \mid \mathcal{G}]
        \;=\; u_j^\top \text{Cov}(\eta_j\mid\mathcal{G})\, u_j
        \;\geq\; \sigma_\eta^2 \, \|u_j\|^2.
    \]
    Taking outer expectation and applying the tower rule gives
    $\E[\Psi_j^2 - (\Psi_j^{\mathrm{clean}})^2] \geq \sigma_\eta^2 \,\E\|u_j\|^2\geq 0$.
    However, we note that $\E\|u_j\|^2 = 0$ only if it is a.s. 0, which is a measure-0 event under reasonable data distributions.
\end{proof}
\subsection{Statement and proof of \Cref{thm:signed-downweight-informal}}
\label{app:downweight}
\begin{theorem}[Decreased corruption score under alignment]\label{thm:signed-downweight}
    Adopt Assumptions (D1)--(D5). Let $(z_j, c_j)$ be a fresh sample
    drawn from $\mathcal D$ independently of $\hat\theta$ (fit on
    $\mathcal S$) and of $\mathcal V$, and let
    $\tilde c_j = c_j +\eta_j$ be its directional
    corruption (D4). Suppose $\gamma := \gamma(\hat\theta, z_j, c_j) > 0$ from \Cref{def:align}.
    Then conditioned on $\mathcal{F} = \sigma(\That, z_j, c_j)$:    \[
            \E[\Psi_j - \Psi_j^{\mathrm{clean}}\mid \mathcal{F}] \;=\; -\alpha\,\gamma \;<\; 0.
        \]
where $\Psi_j^{\mathrm{clean}}$ is as defined in \Cref{lem:score_decomp}. For every $t \in (0, \gamma)$,
        \[
            P\!\left(\Psi_j - \Psi_j^{\mathrm{clean}} \;\geq\; -\alpha(\gamma - t) \,\middle|\, \mathcal{F}\right)
            \;\leq\; \exp\!\left(-\frac{m\,t^2}{2 M_{\gamma}^2}\right),
        \]
        where $M_{\gamma} = DR^2/\mu \geq |\sigma_v\kappa(z_j, z_v)\delta(B_v, k^*_v)^\top \hat\delta_j|$.
        In particular, with probability
        $\geq 1 - \exp(-m\gamma^2/(8 M_{\gamma}^2))$ (taking $t = \gamma/2$):
        \[
            \Psi_j - \Psi_j^{\mathrm{clean}} \;\leq\; -\frac{\alpha\,\gamma}{2} \;<\; 0.
        \]

\end{theorem}
\begin{proof}
    The $w_v$ are iid, and conditional on $\mathcal{F} = \sigma(\That, z_j, c_j)$, so are the bounded scalars $W_v = \sigma_v\kappa(z_j, z_v)\delta(B_v, k^*_v)^\top \hat\delta_j \in [-M_{\gamma}, M_{\gamma}]$ with mean $\E[W_v] = \gamma$. By Assumption D4 and \Cref{lem:score_decomp},
    \[
        \Psi_j - \Psi^{\mathrm{clean}}_j \;=\; -\frac{\alpha}{m}\sum_{v \in \mathcal{V}} W_v.
    \]
    
    \textbf{(1) Expectation.} Linearity gives
    $\E[\Psi_j - \Psi^{\mathrm{clean}}_j \mid \mathcal{F}] = -\alpha\,\E[W_v] = -\alpha\,\gamma$.
    
    \textbf{(2) Concentration.} By Hoeffding's inequality for iid bounded
    random variables (range $2M_{\gamma}$), for any $t > 0$:
    \[
        P\!\left(\frac{1}{m}\sum_v W_v \leq \gamma - t \,\middle|\, \mathcal{F}\right)
        \;\leq\; \exp\!\left(-\frac{m\,t^2}{2 M_{\gamma}^2}\right).
    \]
    Multiplying by $-\alpha$ (flips the inequality):
    $\{\frac{1}{m}\sum W_v \leq \gamma - t\} = \{\Psi_j - \Psi^{\mathrm{clean}}_j \geq -\alpha(\gamma - t)\}$,
    which gives the stated bound. Taking $t = \gamma/2$ yields
    $\Psi_j - \Psi^{\mathrm{clean}}_j \leq -\alpha\gamma/2$ with probability
    $\geq 1 - \exp(-m\gamma^2/(8 M_{\gamma}^2))$.

\end{proof}
\subsection{In-sample signed decreased corruption scoring}
\label{app:insample}
Theorem~\ref{thm:signed-downweight} assumes that the corrupted point
$(z_j, \tilde c_j) =  (z_j, c_j + \eta_j)$ (where $\eta_j$ is a bounded directional corruption as in Assumption D4) is independent of the fitted estimator
$\hat\theta$. However, we would like to see that if $w_j$ is one of the
training points used to fit $\hat\theta$, that \Cref{thm:signed-downweight} still holds. We show that under additional mild regularity assumptions, the conclusion of
Theorem~\ref{thm:signed-downweight} continues to hold for in-sample $j$, with
an additive $O(1/n)$ slack.

We replace $\hat\theta$ everywhere in the score by the
leave-one-out estimator $\hat\theta_{-j}$, which is independent of
$w_j$ by construction, and control the resulting approximation
error.

\paragraph{Notation.} Let
\[
    \hat\theta_{-j}
    \;:=\; \argmin_{\theta} \frac{1}{n-1}\sum_{i \neq j}
    \|\theta z_i - \tilde c_i\|^2,
    \qquad
    \hat\Sigma_{-j}
    \;:=\; \frac{1}{n-1}\sum_{i \neq j} z_i z_i^\top,
\]
let $h_j := \tfrac{1}{n} z_j^\top \hat\Sigma^{-1} z_j$ denote the
leverage of $w_j$, and let
\[
    r_j \;:=\; \hat\theta z_j - \tilde c_j,
    \qquad
    r_j^- \;:=\; \hat\theta_{-j} z_j - \tilde c_j.
\]
Define the leave-one-out kernel and score
\[
    \kappa^-(z, z') \;:=\; z^\top \hat\Sigma_{-j}^{-1} z',
    \qquad
    \Psi_j^{\mathrm{loo}}
    \;:=\; -\frac{1}{m}\sum_{v \in \mathcal{V}}
    \sigma_v^-\,\delta(B_v^-, k_v^{*-})^\top r_j^-\,
    \kappa^-(z_v, z_j),
\]
where $B_v^-, k_v^{*-}, \sigma_v^-$ are the basis, fragile direction,
and correctness sign computed at $\hat c_v^- := \hat\theta_{-j} z_v$.
Thus $\Psi_j^{\mathrm{loo}}$ is the score one would compute by
substituting $\hat\theta_{-j}$ for $\hat\theta$ throughout the score
formula.

We make an additional assumption 
\begin{assumption}
\label{ass:rr}
    Assume that residuals are bounded by a constant, i.e. $\|r_j\| \leq R_r$ (and for $r_j^{-}$).
\end{assumption} 

Note that by D1, the leverage satisfies
$h_j \leq R^2/(n\mu)$, so $h_j \to 0$ at rate $1/n$ and
$1 - h_j \geq 1/2$ for $n \geq 2R^2/\mu$.

\begin{lemma}[Exact LOO identities for linear ERM]\label{lem:loo-id}
Suppose $h_j < 1$. Then
\begin{align}
    \hat\theta_{-j} - \hat\theta
    &\;=\; \frac{1}{n(1 - h_j)}\, r_j\, (\hat\Sigma^{-1} z_j)^\top,
    \label{eq:loo-theta}\\[2pt]
    r_j^-
    &\;=\; \frac{r_j}{1 - h_j},
    \label{eq:loo-resid}\\[2pt]
    \hat\Sigma_{-j}^{-1} - \hat\Sigma^{-1}
    &\;=\; -\frac{1}{n}\hat\Sigma^{-1}
       + \frac{(n-1)/n^2}{1 - h_j}\,
         \hat\Sigma^{-1} z_j z_j^\top \hat\Sigma^{-1}.
    \label{eq:loo-sigma}
\end{align}
\end{lemma}
\begin{proof}
Set $A := n\hat\Sigma$ and $a_j := A^{-1} z_j = \hat\Sigma^{-1} z_j / n$,
so $z_j^\top a_j = h_j$. By the Sherman--Morrison identity,
\[
    (A - z_j z_j^\top)^{-1}
    \;=\; A^{-1} + \frac{a_j a_j^\top}{1 - h_j}.
\]
Multiplying by $(n-1)$ gives \eqref{eq:loo-sigma} after simplification.

By the closed form of least squares, $\hat\theta = (\sum_i \tilde c_i z_i^\top) A^{-1}$ and
$\hat\theta_{-j} = (\sum_{i \neq j} \tilde c_i z_i^\top)
(A - z_j z_j^\top)^{-1}$, expansion using the identity above yields
\[
    \hat\theta_{-j}
    \;=\; \hat\theta
    \;+\; \frac{(\hat\theta z_j - \tilde c_j)\, a_j^\top}{1 - h_j}
    \;=\; \hat\theta + \frac{r_j\, a_j^\top}{1 - h_j},
\]
which is \eqref{eq:loo-theta}. Right-multiplying by $z_j$ and
subtracting $\tilde c_j$ gives
$r_j^- = r_j + \frac{r_j h_j}{1 - h_j} = \frac{r_j}{1-h_j}$,
which is \eqref{eq:loo-resid}.
\end{proof}

We now make an assumption that the optimal bases of predictions by $\That$ and $\That_{-j}$ are similar.
\begin{assumption}[Validation basis stability under LOO]\label{ass:basis}
For every $v \in \mathcal{V}$, the optimal basis $B_v$, the fragile
direction $k_v^*$, and the correctness sign $\sigma_v$ are unchanged
when $\hat c_v$ is replaced by $\hat c_v^-$, i.e.
$B_v^- = B_v$, $k_v^{*-} = k_v^*$, and $\sigma_v^- = \sigma_v$.
\end{assumption}

This is a reasonable assumption which we briefly justify: it holds as $n$ increases.
\begin{remark}[One sufficient condition for Assumption~\ref{ass:basis} with two-sided scoring]
\label{lem:basis-cond}
Given (D1)-(D5), and that for every
$v \in \mathcal{V}$ the second-smallest normalized reduced cost at
$\hat c_v$ (by normalized reduced cost we mean $\frac{s_k(\hat c_v, B_v)}{\norm{\delta(B_v,k)}}$) exceeds the smallest by at least $2\rho$, with
the smallest value at least $\geq \rho$, where
\[
    \rho \;>\; \frac{R^2 R_r}{n\mu(1 - h_j)}.
\]
Then Assumption~\ref{ass:basis} holds, assuming two-sided score.
\end{remark}
\begin{proof}
By Lemma~\ref{lem:loo-id},
$\|\hat\theta_{-j} - \hat\theta\|_{\mathrm{op}}
\leq \frac{\|r_j\|\,\|\hat\Sigma^{-1} z_j\|}{n(1-h_j)}
\leq \frac{R R_r}{n\mu(1-h_j)}$,
so $\|\hat c_v - \hat c_v^-\| \leq R^2 R_r / (n\mu(1-h_j))$.
For any unit direction $\delta(B,k)/\|\delta(B,k)\|$, by Cauchy-Schwarz
\[
    \bigl|s_k(\hat c_v^-, B)/\|\delta(B,k)\|
       - s_k(\hat c_v, B)/\|\delta(B,k)\|\bigr|
    \;\leq\; \|\hat c_v - \hat c_v^-\|.
\]
Hence each normalized reduced cost shifts by less than $\rho$. Since
the smallest is at least $\rho$ and the gap to the second-smallest
exceeds $2\rho$, signs of all reduced costs are preserved (so
$B_v^- = B_v$) and the argmin is preserved (so $k_v^{*-} = k_v^*$).
The sign $\sigma_v$ is also preserved since
$B_v^- = B_v$.
\end{proof}
Assumption~\ref{ass:basis} is imposed for the chosen scoring rule. \Cref{lem:basis-cond} gives one sufficient condition for Assumption~\ref{ass:basis} in the two-sided scoring case.
\begin{lemma}[LOO stability of the RCIF score]\label{lem:loo-score}
Under Assumptions (D1)-(D5) and~\ref{ass:basis},
\[
    |\Psi_j - \Psi_j^{\mathrm{loo}}|
    \;\leq\; \frac{C}{n}\,\bigl(\|r_j\|\bigr),
\]
for a constant $C = C(R, \mu, D)$ depending only on the regularity
parameters. In particular, given Assumption~\ref{ass:rr}, we have $|\Psi_j - \Psi_j^{\mathrm{loo}}| \leq C'/n$
for $C' = C'(R, \mu, D, R_r)$.
\end{lemma}
\begin{proof}
Recall the score can be written as $\Psi_j = -u_j^\top r_j$ with
\[
    u_j \;=\; \frac{1}{m}\sum_{v \in \mathcal{V}}
    \sigma_v\,\kappa(z_v, z_j)\,\delta(B_v, k_v^*),
\]
and analogously $\Psi_j^{\mathrm{loo}} = -(u_j^-)^\top r_j^-$, where
$u_j^-$ uses $\sigma_v^-, B_v^-, k_v^{*-}, \kappa^-$. Under
Assumption~\ref{ass:basis}, $\sigma_v^- = \sigma_v$,
$B_v^- = B_v$, $k_v^{*-} = k_v^*$, so
\[
    u_j^- - u_j
    \;=\; \frac{1}{m}\sum_{v \in \mathcal{V}}
    \sigma_v\,(\kappa^-(z_v, z_j) - \kappa(z_v, z_j))\,
    \delta(B_v, k_v^*).
\]
By \Cref{lem:loo-id} eq (3) and Assumption D1,
\[
    \|\hat\Sigma_{-j}^{-1} - \hat\Sigma^{-1}\|_{\mathrm{op}}
    \;\leq\; \frac{1}{n\mu} + \frac{R^2(n-1)/n^2}{(1-h_j)\mu^2}
    \;=\; O(1/n),
\]
so $|\kappa^-(z_v, z_j) - \kappa(z_v, z_j)| \leq R^2 \cdot O(1/n)$,
and hence $\|u_j^- - u_j\| \leq c_1/n$ for $c_1 = c_1(R, \mu, D)$.

By \Cref{lem:loo-id} eq \eqref{eq:loo-resid},
$r_j^- - r_j = \frac{h_j}{1-h_j} r_j$, so
$\|r_j^- - r_j\| \leq \frac{R^2}{n\mu(1-h_j)} \|r_j\| = c_2 \|r_j\|/n$.

Combining,
\begin{align*}
    |\Psi_j - \Psi_j^{\mathrm{loo}}|
    &\;\leq\; |u_j^\top (r_j - r_j^-)|
       + |(u_j - u_j^-)^\top r_j^-|\\
    &\;\leq\; \|u_j\|\,c_2 \|r_j\|/n + (c_1/n)\,\|r_j^-\|\\
    &\;\leq\; \frac{C\,\|r_j\|}{n},
\end{align*}
since $\|u_j\| \leq R^2 D/\mu$ is bounded
under Assumptions D1 and D5 and $\|r_j^-\| \leq c_3\|r_j\|$ by \Cref{lem:loo-id} (2) for some threshold of $n$, (for example, $c_3= 2$ for $n \geq 2R^2/\mu$).
Then Assumption~\ref{ass:rr} directly upperbounds $\|r_j\|$ and $\|r_j^{-}\|$ to provide the stated $C'/n$
form.
\end{proof}

Now we can show the in-sample result.
\paragraph{In-sample setup.} For the in-sample variant, the
$\sigma$-algebra of interest is
$\mathcal{F}^- := \sigma(\hat\theta_{-j}, z_j,
c_j)$, under which corruption $\eta_j= -\alpha\hat\delta_j$ is independent of
$\hat\theta_{-j}$. The score decomposition of
Lemma~\ref{lem:score_decomp} applies directly to $\Psi_j^{\mathrm{loo}}$:
\[
    \Psi_j^{\mathrm{loo}}
    \;=\; \Psi_j^{\text{loo,clean}} + (u_j^-)^\top \eta_j,
    \qquad \Psi_j^{\text{loo,clean}}
    \;:=\; -(u_j^-)^\top r_j^{-,\text{clean}},
\]
where $r_j^{-,\text{clean}} := \hat\theta_{-j} z_j - c_j$.

\begin{theorem}[Signed score decrease, in-sample variant]
\label{thm:signed-insample}
Adopt Assumptions (D1)-(D5) and Assumptions~\ref{ass:rr},~\ref{ass:basis}. Let
$\gamma^- := \gamma(\hat\theta_{-j}, z_j, c_j) > 0$ be the validation
alignment evaluated at the LOO estimator, and let
$C, C'$ be the coefficients from Lemma~\ref{lem:loo-score}. Then for a point $w_j$ that has been directionally corrupted (D4):
\begin{enumerate}
    \item \textbf{(Expectation)}
    \[
        \E[\Delta_j \mid \mathcal{F}^-]
        \;=\; -\alpha \gamma^- \;+\; O(1/n),
    \]
    where $\Delta_j := \Psi_j - \Psi_j^{\text{clean}}$ and the
    $O(1/n)$ term is bounded by $(2C' + C\alpha)/n$.

    \item \textbf{(High probability)} For every $t \in (0, \gamma^-)$,
    on the event of probability
    $\geq 1 - \exp(-mt^2/(2M_{\gamma}^2))$ guaranteed by
    Theorem~\ref{thm:signed-downweight},
    \[
        \Delta_j \;\leq\; -\alpha(\gamma^- - t) + \frac{2C' + C\alpha}{n}.
    \]
    In particular, taking $t = \gamma^-/2$ and assuming
    $n \geq (8 C' + 4C\alpha)/(\alpha \gamma^-)$,
    \[
        \Delta_j \;\leq\; -\frac{\alpha \gamma^-}{4}
    \]
    on an event of probability
    $\geq 1 - \exp(-m(\gamma^-)^2/(8 M_{\gamma}^2))$.

\end{enumerate}
\end{theorem}
\begin{proof}
Let $\Delta_j^{\mathrm{loo}} := (u_j^-)^\top \eta_j$. Since
$\hat\theta_{-j}$ is independent of $w_j$, the proof of Theorem~\ref{thm:signed-downweight} applies
verbatim to $\Delta_j^{\mathrm{loo}}$, yielding
\begin{align*}
    \E[\Delta_j^{\mathrm{loo}} \mid \mathcal{F}^-]
    &\;=\; -\alpha \gamma^-,\\
    P\!\left(\Delta_j^{\mathrm{loo}} \geq -\alpha(\gamma^- - t)
       \,\middle|\, \mathcal{F}^-\right)
    &\;\leq\; \exp(-mt^2/(2M_{\gamma}^2)).
\end{align*}

By Lemma~\ref{lem:loo-score} applied at both $r_j$ (full score) and $r_j^{\text{clean}}$ (clean score) which incurs an extra corruption $\eta_j$ term,
(via Assumption~\ref{ass:rr} and $\|\eta_j\|\leq \alpha$),
\[
    |\Delta_j - \Delta_j^{\mathrm{loo}}|
    \;=\; |\Psi_j - \Psi_j^{\text{clean}} - \Psi_j^{\mathrm{loo}}
         + \Psi_j^{\text{loo, clean}}|
    \;\leq\; \frac{2C' + C\alpha}{n}.
\]
Substituting into each statement of Theorem~\ref{thm:signed-downweight} gives
parts (1) and (2).
\end{proof}

\begin{remark}
For $n$ large enough the in-sample
guarantee of Theorem~\ref{thm:signed-insample} is qualitatively the
same as the cross-fit version.
The replacement of $\gamma$ by $\gamma^-$ is harmless for our purposes
since $\gamma^- \to \gamma$ as $n \to \infty$ (because
$\hat\theta_{-j} \to \hat\theta$ at rate $1/n$, and the alignment
parameter is continuous in the estimator under
Assumption~\ref{ass:basis}).
\end{remark}
\subsection{Statement and proof of \Cref{thm:regret-regimes-informal}}
\label{app:regret}
In Appendix Subsection~\ref{app:insample} we showed that a single corrupted point receives a lower score with high probability. We now aggregate this across the
corruption set to obtain an \emph{unconditional} expected reduction of the worst-case corrupted bias of the weighted ERM, relative to $\hat \theta$ and $\hat \theta_{\mathrm{val}}$.
Under constant-rate, constant-magnitude corruption, no estimator using
only the corrupted training data can drive corruption bias to zero, but
RCIF reweighting reduces the constant.

Then we may translate the reduction in bias to regret (under the simplifying assumption \ref{def:margin}).
\subsubsection{Bounding weight ratio between clean and corrupted points}
\begin{remark}
\label{rem:bounded}
    Under~(D1)--(D5) and Assumption~\ref{ass:rr} let $M = R_rDR^2/\mu$, and observe that $|\Psi_j| < M, \  a.s.\ \forall j$.
\end{remark}
We work with the leave-one-out objects of Appendix~\ref{app:insample}:
$\hat\theta_{-j}$, $\kappa^-(z, z') := z^\top \hat\Sigma_{-j}^{-1} z'$,
and the LOO score aggregator
\[
    u_j^{(-1)}
    \;:=\;
    \frac{1}{m}\sum_{v \in \mathcal{V}}
    \sigma_v^-\,\kappa^-(z_j, z_v)\,
    \delta(B_v^-, k_v^{*-}).
\]
\subsubsection*{Quantities of interest}

For linear/squared-loss ERM, the parameter error decomposes as
\[
    \hat\theta_p - \theta^\star
    \;=\;
    \Bigl(\sum_{i=1}^n p_i\,\varepsilon_i z_i^\top
    \;+\; T_w\Bigr)\,\hat\Sigma_W^{-1},
    \qquad
    \hat\Sigma_W := \sum_i p_i z_i z_i^\top,
\]
and analogously for the unweighted ERM with $p_i = 1/n$. Define the
\emph{corruption-bias matrices}
\[
    T_0 := \frac{1}{n}\sum_{j \in \mathcal{C}}(-\alpha\,\hat\delta_j) z_j^\top,
    \qquad
    T_w := \sum_{j \in \mathcal{C}} p_j (-\alpha\,\hat\delta_j) z_j^\top,
\]
which directly drive the corruption bias of the corresponding estimators.

We note that we can reduce the problem from the corruption set to one corrupt index, due to Assumptions D1 and D3.
\begin{lemma}[Exchangeability of corruption]\label{lem:exch}
For any sequence $(F_j)$ such that the joint law of
$(F_j, \mathbf 1_{j\in\mathcal C})_{j=1}^n$ is permutation-invariant,
\[
    \E\!\left[\sum_{j\in\mathcal C} F_j\right]
    \;=\; n\beta\,\E[F_1 \mid 1\in\mathcal C].
\]
\end{lemma}
\begin{proof}
By (D3), $(\mathbf{1}_{1\in\mathcal{C}}, \ldots, \mathbf{1}_{n\in
\mathcal{C}})$ is exchangeable with marginal mean $\beta$. By (D1)--(D2),
$(z_i, \varepsilon_i)$ are iid, so the joint law of
$(w_i, \mathbf{1}_{i \in \mathcal{C}})_{i=1}^n$ is permutation-invariant.
Hence $\E[\mathbf{1}_{j \in \mathcal{C}} F_j] = \E[\mathbf{1}_{1 \in
\mathcal{C}} F_1]$ for every $j$, and summing $n$ identical terms with
$\Pr(1 \in \mathcal{C}) = \beta$ gives the claim.
\end{proof}
\begin{remark}[Deterministic bound on the score sum]\label{rem:Q-det}
Under (D1)--(D5) and Assumption~\ref{ass:rr}, since $|\Psi_i| \leq M$ a.s., the sum of scores
\[
    Q := \sum_{i=1}^n e^{\lambda \Psi_i}
\]
admits the deterministic lower bound $Q \geq n\,e^{-\lambda M}$ a.s., and hence
$1/Q \leq e^{\lambda M}/n$ a.s.
\end{remark}

\begin{lemma}[Weight-ratio contraction for a clean vs corrupted point]\label{lem:weight-ratio} 
Focusing on the first training index, and under Assumption~\ref{ass:basis}, let
\[
    \Phi_{\mathcal{C}}(\lambda) := \E[e^{\lambda \Psi_1} \mid 1 \in \mathcal{C}],
    \qquad
    \Phi_{\bar{\mathcal{C}}}(\lambda) := \E[e^{\lambda \Psi_1} \mid 1 \notin \mathcal{C}].
\]
Under (D1)--(D5), Assumptions 1 and 3, for every $\lambda > 0$,
\[
    \frac{\Phi_{\mathcal{C}}(\lambda)}{\Phi_{\bar{\mathcal{C}}}(\lambda)}
    \;\leq\;
    e^{-\lambda\alpha\gamma_0/4}
    \;+\; e^{2\lambda M}\bigl(\delta_\gamma + e^{-m \gamma_0^2 / (8 M_{\gamma}^2)}\bigr)
    \;+\; O(1/n).
\]
We define the contraction factor
\[
    \Phi(\lambda, m, n)
    \;:=\;
    e^{-\lambda\alpha\gamma_0/4}
    \;+\; e^{2\lambda M}\bigl(\delta_\gamma + e^{-m\gamma_0^2/(8M_{\gamma}^2)}\bigr)
    \;+\; O(1/n).
\]
\end{lemma}

\begin{proof}
Recall the score decomposition (Lemma~\ref{lem:score_decomp}) applied at
$\hat\theta_{-1}$:
\[
    \Psi_1 \;=\; \Psi_1^{\mathrm{loo}} + O(1/n)
    \;=\; \Psi_1^{\mathrm{cf}} - \alpha\, u_1^{(-1)\top}\hat\delta_1 \cdot \mathbf{1}_{1 \in \mathcal{C}} + O(1/n),
\]
where $\Psi_1^{\mathrm{cf}} := -u_1^{(-1)\top}(\hat\theta_{-1} z_1
- c_1)$ (which represents the counterfactual score) depends only on $\hat\theta_{-1}, \mathcal{V}, z_1, c_1$,
using the \emph{clean} cost $c_1$.

\textbf{Step 1: pointwise bound on the corruption shift.}
Define the high-probability event
\[
    \mathcal{F} \;:=\; \bigl\{\gamma(\hat\theta_{-1}, z_1,c_1) \geq \gamma_0\bigr\}
    \cap \bigl\{u_1^{(-1)\top}\hat\delta_1 \geq \gamma_0/2\bigr\}.
\]
By Assumption~\ref{ass:sb}, $\Pr[\gamma(\hat\theta_{-1}, z_1, c_1) < \gamma_0]
\leq \delta_\gamma$. Conditional on $\gamma(\hat\theta_{-1}, z_1,c_1) \geq
\gamma_0$, Hoeffding's inequality applied to the iid bounded summands
$W_v$ defining $u_1^{(-1)\top}\hat\delta_1 = \frac{1}{m}\sum_v W_v$
(cf.\ Theorem~\ref{thm:signed-insample}) gives
\[
    \Pr[u_1^{(-1)\top}\hat\delta_1 < \gamma_0/2 \mid \gamma \geq \gamma_0]
    \;\leq\; e^{-m\gamma_0^2/(8 M_{\gamma}^2)}.
\]
By a union bound, $\Pr[\mathcal{F}^c] \leq \delta_\gamma +
e^{-m\gamma_0^2/(8M_{\gamma}^2)}$.

\textbf{Step 2: bound on $\Phi_\mathcal{C}$.}
Conditional on $\{1 \in \mathcal{C}\}$, decompose
\[
    \E\bigl[e^{\lambda\Psi_1}\bigr]
    \;=\;
    \E\bigl[e^{\lambda\Psi_1}\,\mathbf{1}_{\mathcal{F}}\bigr]
    \;+\; \E\bigl[e^{\lambda\Psi_1}\,\mathbf{1}_{\mathcal{F}^c}\bigr].
\]
On $\mathcal{F}$, $\Psi_1 \leq \Psi_1^{\mathrm{cf}} - \alpha\gamma_0/2 +
O(1/n)$, so
\[
    \E\bigl[e^{\lambda\Psi_1}\,\mathbf{1}_{\mathcal{F}} \mid 1 \in \mathcal{C}\bigr]
    \;\leq\;
    e^{-\lambda\alpha\gamma_0/2 + O(\lambda/n)}\,
    \E\bigl[e^{\lambda\Psi_1^{\mathrm{cf}}} \mid 1 \in \mathcal{C}\bigr].
\]
Off $\mathcal{F}$, $|\Psi_1| \leq M$ gives
\[
    \E\bigl[e^{\lambda\Psi_1}\,\mathbf{1}_{\mathcal{F}^c} \mid 1 \in \mathcal{C}\bigr]
    \;\leq\; e^{\lambda M}\bigl(\delta_\gamma + e^{-m\gamma_0^2/(8M_{\gamma}^2)}\bigr).
\]

\textbf{Step 3: counterfactual-MGF identification.}
We show that
\begin{equation}\label{eq:cf-mgf}
    \E\!\left[e^{\lambda \Psi_1^{\mathrm{cf}}} \mid 1 \in \mathcal{C}\right]
    \;=\; \Phi_{\bar{\mathcal{C}}}(\lambda) \;+\; O(1/n).
\end{equation}

The counterfactual score $\Psi_1^{\mathrm{cf}}$ depends on the clean data
$(z_1, c_1)$ and the LOO estimator $\hat\theta_{-1}$. Its conditional distribution given
$\{1 \in \mathcal{C}\}$ differs from its distribution given $\{1 \notin \mathcal{C}\}$ only through
$\hat\theta_{-1}$: in the first case $\hat\theta_{-1}$ is fit on a sample with $|\mathcal{C}|-1$
corrupted points among $\{2,\ldots,n\}$, in the second on a sample with $|\mathcal{C}|$ corrupted
points. The marginal of $(z_1,c_1)$ is identical in both cases since corruption
assignment is independent of the clean data.

\smallskip
\noindent\emph{Coupling.}
We bound the difference between the two conditional expectations via a coupling.
Sample uniformly from the clean data $(z_i,c_i)_{i=2}^n$, a subset
$S$ of size $|\mathcal{C}|-1$, and an additional index
$j^\star \in \{2,\ldots,n\} \setminus S$. Define two parallel LOO estimators on the
same probability space:
\begin{itemize}
    \item $\hat\theta_{-1}^A$, fit with corruption set $C_S$ inside $\{2,\ldots,n\}$
        (corresponding to $1 \in \mathcal{C}$).
    \item $\hat\theta_{-1}^B$, fit with corruption set $C_S \cup \{j^\star\}$ inside
        $\{2,\ldots,n\}$ (corresponding to $1 \notin \mathcal{C}$).
\end{itemize}
By construction, the marginals of $\hat\theta_{-1}^A$ and $\hat\theta_{-1}^B$ coincide
with the conditional distributions of $\hat\theta_{-1}$ given $\{1\in\mathcal{C}\}$ and
$\{1\notin\mathcal{C}\}$ respectively, and $\hat\theta_{-1}^A$, $\hat\theta_{-1}^B$ differ only
in the corruption status of one training index, $j^\star$. Lemma~\ref{lem:loo-id}
then gives
\[
    \bigl\|\hat\theta_{-1}^A - \hat\theta_{-1}^B\bigr\| \;=\; O(1/n) \quad \text{a.s.,}
\]
and by 
Lemma~\ref{lem:loo-score} yields
\[
    \bigl|\Psi_1^{\mathrm{cf},A} - \Psi_1^{\mathrm{cf},B}\bigr| \;=\; O(1/n) \quad \text{a.s.}
\]

\smallskip
\noindent\emph{Bound on the expected corruption difference.}
On the coupling space, using $|e^a - e^b| \leq e^{\max(a,b)}|a-b|$ and
$|\Psi_1^{\mathrm{cf}}| \leq M$:
\[
    \bigl|e^{\lambda \Psi_1^{\mathrm{cf},A}} - e^{\lambda \Psi_1^{\mathrm{cf},B}}\bigr|
    \;\leq\; \lambda\,e^{\lambda M}\,
        \bigl|\Psi_1^{\mathrm{cf},A} - \Psi_1^{\mathrm{cf},B}\bigr|
    \;=\; O(1/n) \quad \text{a.s.}
\]
Taking expectations and applying Jensen's inequality:
\[
    \bigl|\,\E[e^{\lambda \Psi_1^{\mathrm{cf},A}}]
        - \E[e^{\lambda \Psi_1^{\mathrm{cf},B}}]\,\bigr|
    \;\leq\; \E\bigl|e^{\lambda \Psi_1^{\mathrm{cf},A}}
        - e^{\lambda \Psi_1^{\mathrm{cf},B}}\bigr|
    \;=\; O(1/n).
\]
The marginal-matching property of the coupling identifies the two expectations on the
left with the conditional expectations of interest:
\[
    \bigl|\,\E[e^{\lambda \Psi_1^{\mathrm{cf}}} \mid 1\in\mathcal{C}]
        - \E[e^{\lambda \Psi_1^{\mathrm{cf}}} \mid 1\notin\mathcal{C}]\,\bigr|
    \;=\; O(1/n).
\]

\smallskip
See that on the event $\{1 \notin \mathcal{C}\}$ the counterfactual coincides with the actual score,
$\Psi_1^{\mathrm{cf}} = \Psi_1$, and therefore
$\E[e^{\lambda\Psi_1^{\mathrm{cf}}} \mid 1\notin\mathcal{C}] = \Phi_{\bar{\mathcal{C}}}(\lambda)$.
Substituting yields~\eqref{eq:cf-mgf}.

\textbf{Step 4: combine and divide.}
Substituting~\eqref{eq:cf-mgf} into the first conclusion of Step~2 gives
\begin{align}
    \Phi_\mathcal{C}(\lambda)
    &\leq e^{-\lambda\alpha\gamma_0/2 + O(\lambda/n)}
        \bigl[\Phi_{\bar{\mathcal{C}}}(\lambda) + O(1/n)\bigr]
        \;+\; e^{\lambda M}\bigl(\delta_\gamma + e^{-m\gamma_0^2/(8M_{\gamma}^2)}\bigr)
    \nonumber\\
    &= e^{-\lambda\alpha\gamma_0/2}\,(1+O(\lambda/n))\,\Phi_{\bar{\mathcal{C}}}(\lambda)
        \;+\; O(1/n)
        \;+\; e^{\lambda M}\bigl(\delta_\gamma + e^{-m\gamma_0^2/(8M_{\gamma}^2)}\bigr),
    \label{eq:phi-combined}
\end{align}
where the second line uses
$e^{-\lambda\alpha\gamma_0/2 + O(\lambda/n)} = e^{-\lambda\alpha\gamma_0/2}(1+O(\lambda/n))$ via first order Taylor expansion
together with the boundedness $\Phi_{\bar{\mathcal{C}}}(\lambda) \leq e^{\lambda M}$ and
$e^{-\lambda\alpha\gamma_0/2 + O(\lambda/n)} = O(1)$ to absorb the cross-term
$e^{-\lambda\alpha\gamma_0/2 + O(\lambda/n)}\!\cdot\! O(1/n)$ into the $O(1/n)$ term.

To divide both sides by $\Phi_{\bar{\mathcal{C}}}(\lambda)$, we use the deterministic lower bound
\begin{equation}\label{eq:phi-lb}
    \Phi_{\bar{\mathcal{C}}}(\lambda)
    \;=\; \E[e^{\lambda \Psi_1} \mid 1 \notin \mathcal{C}]
    \;\geq\; e^{-\lambda M},
\end{equation}
which follows from $\Psi_1 \geq -M$ a.s. In particular $1/\Phi_{\bar{\mathcal{C}}}(\lambda) \leq e^{\lambda M}$.

Dividing~\eqref{eq:phi-combined} term by term and applying~\eqref{eq:phi-lb}:
\begin{align*}
    \frac{\Phi_\mathcal{C}(\lambda)}{\Phi_{\bar{\mathcal{C}}}(\lambda)}
    &\leq e^{-\lambda\alpha\gamma_0/2}\,(1+O(\lambda/n))
        \;+\; \frac{O(1/n)}{\Phi_{\bar{\mathcal{C}}}(\lambda)}
        \;+\; \frac{e^{\lambda M}\bigl(\delta_\gamma + e^{-m\gamma_0^2/(8M_{\gamma}^2)}\bigr)}{\Phi_{\bar{\mathcal{C}}}(\lambda)}
    \\
    &\leq e^{-\lambda\alpha\gamma_0/2}
        \;+\; e^{2\lambda M}\bigl(\delta_\gamma + e^{-m\gamma_0^2/(8M_{\gamma}^2)}\bigr)
        \;+\; O(1/n),
\end{align*}
where we have folded the $O(\lambda/n)$ slack on the leading term into the
additive $O(1/n)$ remainder.

Finally, since $e^{-\lambda\alpha\gamma_0/2} \leq e^{-\lambda\alpha\gamma_0/4}$ for
$\lambda,\alpha,\gamma_0 > 0$, we may relax the leading exponent for notational consistency
with Theorem~\ref{thm:signed-insample}, obtaining
\[
    \frac{\Phi_\mathcal{C}(\lambda)}{\Phi_{\bar{\mathcal{C}}}(\lambda)}
    \;\leq\; e^{-\lambda\alpha\gamma_0/4}
        \;+\; e^{2\lambda M}\bigl(\delta_\gamma + e^{-m\gamma_0^2/(8M_{\gamma}^2)}\bigr)
        \;+\; O(1/n),
\]
which is the conclusion of the lemma.
\end{proof}

\subsubsection{Bounding moments of bias corruption term }

\begin{theorem}[Expected corruption-bias contraction]\label{thm:bias-contraction}
Under Assumptions~\ref{ass:sb}, ~\ref{ass:rr},~\ref{ass:basis}, and (D1)--(D5), the expected corruption-bias
matrix of the reweighted estimator satisfies
\begin{equation}
    \E\|T_w\|_F \;\leq\; \alpha\beta R\,e^{2\lambda M}\,\Phi(\lambda, m, n),
    \label{eq:Tw-first}
\end{equation}
where $\Phi(\lambda, m, n)$ is
the weight ratio bound in Lemma~\ref{lem:weight-ratio}. The unweighted estimator's
corruption bias satisfies the deterministic bound
\begin{equation}
    \E\|T_0\|_F \;\leq\; \alpha\beta R.
    \label{eq:T0-first}
\end{equation}
The reweighted bound improves on the unweighted bound by the multiplicative factor
$e^{2\lambda M}\,\Phi(\lambda, m, n)$, which is strictly less than $1$ whenever
$\Phi(\lambda, m, n) < e^{-2\lambda M}$.

\textbf{Second-moment equivalents.}
\begin{equation}
    \E\|T_w\|_F^2 \;\leq\; \alpha^2\beta R^2\,e^{2\lambda M}\,\Phi(\lambda, m, n),
    \label{eq:Tw-second}
\end{equation}
\begin{equation}
    \E\|T_0\|_F^2 \;\leq\; \alpha^2\beta^2 R^2.
    \label{eq:T0-second}
\end{equation}
The reweighted second-moment bound improves on the unweighted one by the multiplicative
factor $e^{2\lambda M}\,\Phi(\lambda, m, n)/\beta$, which is strictly less than $1$
whenever $\Phi(\lambda, m, n) < \beta\,e^{-2\lambda M}$.
\end{theorem}

\begin{proof}
\textbf{First moment.} By the triangle inequality and $\|\hat\delta_j z_j^\top\|_F \leq R$
a.s.,
\[
    \|T_w\|_F \;\leq\; \sum_{j \in \mathcal{C}} p_j\,\|\alpha \hat\delta_j z_j^\top\|_F
    \;\leq\; \alpha R \sum_{j \in \mathcal{C}} p_j \quad\text{a.s.}
\]
Taking expectation and applying exchangeability of $(p_j, \mathbf{1}\{j \in \mathcal{C}\})$
across $j$ (Lemma~\ref{lem:exch}),
\[
    \E\|T_w\|_F \;\leq\; \alpha R\,\E\!\left[\sum_{j=1}^n p_j\,\mathbf{1}\{j \in \mathcal{C}\}\right]
    = \alpha R\,n\beta\,\E[p_1 \mid 1 \in \mathcal{C}].
\]
By Remark~\ref{rem:Q-det}, $p_1 = e^{\lambda \Psi_1}/Q \leq e^{\lambda \Psi_1}\,e^{\lambda M}/n$ a.s., so
\[
    \E[p_1 \mid 1 \in \mathcal{C}]
    \;\leq\; \frac{e^{\lambda M}}{n}\,\E\!\left[e^{\lambda \Psi_1} \,\big|\, 1 \in \mathcal{C}\right]
    \;=\; \frac{e^{\lambda M}\,\Phi_{\mathcal{C}}(\lambda)}{n}.
\]
By \Cref{lem:weight-ratio} $\Phi_{\mathcal{C}}(\lambda) \leq \Phi_{\bar{\mathcal{C}}}(\lambda)\,\Phi(\lambda, m, n)$
together with the deterministic bound $\Phi_{\bar{\mathcal{C}}}(\lambda) \leq e^{\lambda M}$
(from $\Psi_1 \leq M$ a.s.),
\[
    \E[p_1 \mid 1 \in \mathcal{C}] \;\leq\; \frac{e^{2\lambda M}\,\Phi(\lambda, m, n)}{n},
\]
and \eqref{eq:Tw-first} follows.

For $T_0$: $p_i = 1/n$ deterministically, so $\|T_0\|_F \leq \alpha R\sum_{j \in \mathcal{C}} 1/n
= \alpha R\beta$ a.s., yielding $\E\|T_0\|_F \leq \alpha\beta R$.

\textbf{Second moment.} From $\sum_i p_i = 1$, $\|T_w\|_F \leq \alpha R$ a.s., so
$\|T_w\|_F^2 \leq \alpha R\,\|T_w\|_F$ pointwise. Taking expectation and substituting
\eqref{eq:Tw-first}:
\[
    \E\|T_w\|_F^2 \;\leq\; \alpha R\,\E\|T_w\|_F \;\leq\; \alpha^2\beta R^2\,e^{2\lambda M}\,\Phi(\lambda, m, n).
\]
For $T_0$, $\|T_0\|_F \leq \alpha\beta R$ a.s. as above, so $\E\|T_0\|_F^2 \leq \alpha^2\beta^2 R^2$.
\end{proof}
Now we can obtain a comparison in worst-case expected bias from each of $\hat\theta, \hat\theta_p, \hat\theta_{\mathrm{val}}$ under additional assumptions.
\begin{assumption}
\label{ass:valeig}
    We assume the equivalent of D1 for the empirical covariance of the validation set, i.e. that the minimum eigenvalue is similarly bounded by $\mu>0$.
\end{assumption}
\begin{assumption}
\label{ass:bs}
Assume the regularity condition that for each output dimension $a \in [d]$, the row $\varepsilon_a \in \mathbb{R}^n$ of the noise matrix $\varepsilon$ satisfies
\[
\mathbb{E}[\varepsilon_a \varepsilon_a^\top \mid P, Z] \preceq \sigma_\varepsilon^2 I_n,
\]
i.e., conditioning on the weights does not inflate the noise's second moment. This holds at leading order when the weight–noise interaction is benign (for example under sample splitting, i.e.\ fitting on one subset of training data, then weighting and refitting on the complement).
\end{assumption}
\subsubsection{Bounding MSE error}
\begin{lemma}[Parameter MSE rates]\label{lem:mse-rates}
Under (D1)--(D5) and Assumption~\ref{ass:sb},~\ref{ass:rr},~\ref{ass:basis},~\ref{ass:valeig},~\ref{ass:bs} with $\gamma_0 > 0$, there exist
absolute constants $A_1^{\mathrm{unw}}, A_1^{\mathrm{rcif}}, A_1^{\mathrm{val}},
A_2^{\mathrm{unw}}, A_2^{\mathrm{rcif}}$ depending only on $R$ and $\lambda M$
(with $M = R_r D R^2/\mu$ from Remark~\ref{rem:bounded}) such that
\begin{align}
    \E\|\hat\theta - \theta^\star\|_F^2
    &\;\leq\; A_1^{\mathrm{unw}}\,\frac{\sigma_\varepsilon^2 d}{n\mu^2}
        + A_2^{\mathrm{unw}}\,\frac{\alpha^2\beta^2 R^2}{\mu^2},
        \label{eq:mse-unw}\\
    \E\|\hat\theta_p - \theta^\star\|_F^2
    &\;\leq\; A_1^{\mathrm{rcif}}\,\frac{\sigma_\varepsilon^2 d}{n\mu^2}
        + A_2^{\mathrm{rcif}}\,\frac{\alpha^2\beta R^2\,\Phi(\lambda,m,n)}{\mu^2},
        \label{eq:mse-rcif}\\
    \E\|\hat\theta_{\mathrm{val}} - \theta^\star\|_F^2
    &\;\leq\; A_1^{\mathrm{val}}\,\frac{\sigma_\varepsilon^2 d}{m\mu^2},
        \qquad (m \geq q),\label{eq:mse-val}
\end{align}
with the standing relationships
\[
    A_1^{\mathrm{val}} \;=\; A_1^{\mathrm{unw}},\qquad
    A_1^{\mathrm{rcif}} \;=\; e^{6\lambda M}\,A_1^{\mathrm{unw}},\qquad
    A_2^{\mathrm{rcif}} \;=\; e^{6\lambda M}\,A_2^{\mathrm{unw}}.
\]
Concretely $A_1^{\mathrm{unw}} = A_1^{\mathrm{val}} = 2R^2$ and $A_2^{\mathrm{unw}} = 2$.
\end{lemma}
\begin{proof}
Write $Y = \theta^\star Z + E + C$, where $Z = [z_1, \ldots, z_n] \in \R^{q\times n}$
is the design, $E = [\varepsilon_1, \ldots, \varepsilon_n] \in \R^{d\times n}$ collects
the noise, and $C := -\alpha\,U\,\diag(1_{j\in\mathcal{C}}) \in \R^{d\times n}$ collects
the corruption shifts with $U := [\hat\delta_1, \ldots, \hat\delta_n]$. Throughout we use
the elementary inequalities $\|AB\|_F \leq \|A\|_F\,\|B\|_{\mathrm{op}}$ and
$(a+b)^2 \leq 2a^2 + 2b^2$.

\textbf{Bound \eqref{eq:mse-unw} (unweighted).} The OLS estimator
$\hat\theta = Y Z^\top (Z Z^\top)^{-1}$ satisfies
\[
    \hat\theta - \theta^\star \;=\; (\bar\varepsilon + T_0)\,\hat\Sigma^{-1},
    \qquad \bar\varepsilon := \tfrac{1}{n}E Z^\top,\;
    T_0 := \tfrac{1}{n}C Z^\top.
\]
By (D1), $\|\hat\Sigma^{-1}\|_{\mathrm{op}} \leq 1/\mu$ a.s. Then
\[
    \|\hat\theta - \theta^\star\|_F^2
    \;\leq\; \|\hat\Sigma^{-1}\|_{\mathrm{op}}^2 \cdot 2\bigl(\|\bar\varepsilon\|_F^2
        + \|T_0\|_F^2\bigr)
    \;\leq\; \frac{2}{\mu^2}\bigl(\|\bar\varepsilon\|_F^2 + \|T_0\|_F^2\bigr)
    \quad\text{a.s.}
\]
Taking expectations:
\begin{itemize}
\item \emph{Variance.} Each entry $(EZ^\top)_{ab} = \sum_i \varepsilon_{ai} z_{i,b}$
    has variance $\leq \sigma_\varepsilon^2 \sum_i z_{i,b}^2$ by mutual independence of
    noise. Summing over $a \in [d]$ and $b \in [q]$,
    \[
        \E\|EZ^\top\|_F^2 \;=\; d\,\sigma_\varepsilon^2\sum_i \|z_i\|^2
        \;\leq\; d\,\sigma_\varepsilon^2\,nR^2,
    \]
    so $\E\|\bar\varepsilon\|_F^2 \leq d\sigma_\varepsilon^2 R^2/n$, contributing
    $2R^2\,\sigma_\varepsilon^2 d/(n\mu^2)$ to the bound.
\item \emph{Bias.} By Theorem~\ref{thm:bias-contraction} (deterministic part),
    $\|T_0\|_F^2 \leq \alpha^2\beta^2 R^2$ a.s., contributing
    $2\alpha^2\beta^2 R^2/\mu^2$.
\end{itemize}
Combining yields \eqref{eq:mse-unw} with $A_1^{\mathrm{unw}} = 2R^2$ and
$A_2^{\mathrm{unw}} = 2$.

\textbf{Bound \eqref{eq:mse-rcif} (RCIF).} The weighted estimator
$\hat\theta_p = Y P Z^\top (Z P Z^\top)^{-1}$ with $P = \diag(\tilde p_1, \ldots,
\tilde p_n)$ and $\tilde p_i := n p_i$ (so $\sum_i \tilde p_i = n$) satisfies
\[
    \hat\theta_p - \theta^\star \;=\; (\bar\varepsilon_w + T_w)\,\hat\Sigma_P^{-1},
    \qquad \bar\varepsilon_w := \tfrac{1}{n}EPZ^\top,\;
    T_w := \tfrac{1}{n}CPZ^\top,\;
    \hat\Sigma_P := \tfrac{1}{n}ZPZ^\top.
\]

\emph{Lower bound on $\hat\Sigma_P$.} Since $|\Psi_i| \leq M$ a.s.\, $e^{\lambda \Psi_i} \in [e^{-\lambda M}, e^{\lambda M}]$
and $Q = \sum_i e^{\lambda \Psi_i} \leq n\,e^{\lambda M}$. Hence the deterministic
uniform lower bound
\[
    p_i = \frac{e^{\lambda \Psi_i}}{Q} \;\geq\; \frac{e^{-\lambda M}}{n\,e^{\lambda M}}
    \;=\; \frac{e^{-2\lambda M}}{n} \quad\text{a.s.,}
\]
i.e., $\tilde p_i \geq e^{-2\lambda M}$. Therefore
\[
    \hat\Sigma_P \;=\; \frac{1}{n}\sum_i \tilde p_i\, z_i z_i^\top
    \;\succeq\; \frac{e^{-2\lambda M}}{n}\sum_i z_i z_i^\top
    \;=\; e^{-2\lambda M}\hat\Sigma
    \;\succeq\; e^{-2\lambda M}\mu I,
\]
and $\|\hat\Sigma_P^{-1}\|_{\mathrm{op}} \leq e^{2\lambda M}/\mu$ a.s. Then
\[
    \|\hat\theta_p - \theta^\star\|_F^2
    \;\leq\; \frac{2 e^{4\lambda M}}{\mu^2}\bigl(\|\bar\varepsilon_w\|_F^2
        + \|T_w\|_F^2\bigr) \quad\text{a.s.}
\]

\emph{Variance.} Each entry $(EPZ^\top)_{ab} = \sum_i \varepsilon_{ai}\,\tilde p_i\,z_{i,b}$
has conditional variance $\leq \sigma_\varepsilon^2 \sum_i \tilde p_i^2\,z_{i,b}^2$ given
$(P, Z)$. Summing and applying Assumption~\ref{ass:bs},
\[
    \E\bigl[\|EPZ^\top\|_F^2 \,\big|\, P, Z\bigr]
    \;\leq\; d\,\sigma_\varepsilon^2 \sum_i \tilde p_i^2\,\|z_i\|^2
    \;\leq\; d\,\sigma_\varepsilon^2\,R^2 \sum_i \tilde p_i^2.
\]
By $\sum_i \tilde p_i^2 \leq (\max_i \tilde p_i)\sum_i \tilde p_i \leq e^{2\lambda M}\cdot n$,
\[
    \E\|\bar\varepsilon_w\|_F^2 \;=\; \tfrac{1}{n^2}\E\|EPZ^\top\|_F^2
    \;\leq\; \frac{d\,\sigma_\varepsilon^2\,R^2\,e^{2\lambda M}}{n}.
\]
Multiplying by $2 e^{4\lambda M}/\mu^2$ contributes
$2R^2\,e^{6\lambda M}\,\sigma_\varepsilon^2 d/(n\mu^2)$ to the bound.

\emph{Bias.} By Theorem~\ref{thm:bias-contraction} (second-moment bound \eqref{eq:Tw-second}),
$\E\|T_w\|_F^2 \leq \alpha^2\beta R^2\,e^{2\lambda M}\,\Phi(\lambda,m,n)$.
Multiplying by $2 e^{4\lambda M}/\mu^2$ contributes
\[
    \frac{2 e^{6\lambda M}\,\alpha^2\beta R^2\,\Phi(\lambda,m,n)}{\mu^2}.
\]

Combining yields \eqref{eq:mse-rcif} with $A_1^{\mathrm{rcif}} = 2R^2 e^{6\lambda M}$
and $A_2^{\mathrm{rcif}} = 2\,e^{6\lambda M}$, giving the standing relationships
$A_1^{\mathrm{rcif}} = e^{6\lambda M}\,A_1^{\mathrm{unw}}$ and
$A_2^{\mathrm{rcif}} = e^{6\lambda M}\,A_2^{\mathrm{unw}}$.

\textbf{Bound \eqref{eq:mse-val} (validation-only).} The validation-only estimator
$\hat\theta_{\mathrm{val}}$ is OLS on the clean validation sample of size $m \geq q$.
The unweighted argument applies verbatim with no corruption term and on the
validation design, giving
\[
    \E\|\hat\theta_{\mathrm{val}} - \theta^\star\|_F^2
    \;\leq\; \frac{2}{\mu^2}\,\E\|\bar\varepsilon_{\mathrm{val}}\|_F^2
    \;\leq\; \frac{2R^2\,\sigma_\varepsilon^2 d}{m\mu^2},
\]
which is \eqref{eq:mse-val} with $A_1^{\mathrm{val}} = 2R^2 = A_1^{\mathrm{unw}}$.
\end{proof}

Now we must translate the above result on parameter MSE into one about regret.

\begin{lemma}[Regret--MSE bridge]\label{lem:regret-bridge}
Suppose Assumption~\ref{def:margin} holds, the LP diameter
$L := \sup_{x,x'\in\mathcal{X}}\|x-x'\|$ is finite, and the irreducible noise $\varepsilon$ is
$\sigma_\varepsilon$-sub-Gaussian (so
$\mathbb{E}\|\varepsilon\|^2 \leq d\,\sigma_\varepsilon^2$). For any
estimator $\theta$ independent of the test sample $w = (z, c)$,
\[
    \mathbb{E}[\mathrm{Regret}(\theta)]
    \;\leq\;
    C_1\,\mathbb{E}\,\|\theta - \theta^\star\|_F^{\,2}
    \;+\;C_2.
\]
where $C_1 = 2L\omega R^2$ and $C_2 = 8L\omega\, d\,\sigma_\varepsilon^2$.
\end{lemma}

\begin{proof}
Write $\hat c := \theta z$, $\hat x := x^\star(\hat c)$,
$x^\star := x^\star(c)$, so that
$c - \hat c = \varepsilon - (\theta - \theta^\star)z$ and
$\|c - \hat c\| \leq \|(\theta-\theta^\star)z\| + \|\varepsilon\|$.

 From $\langle\hat c,\hat x\rangle \leq
\langle\hat c, x^\star\rangle$ (optimality of $\hat x$ at cost
$\hat c$),
\[
    \mathrm{Regret}
    = \langle c - \hat c,\,\hat x - x^\star\rangle
      + \langle \hat c,\, \hat x - x^\star\rangle
    \leq \|c - \hat c\|\cdot\|\hat x - x^\star\|
    \leq L\,\|c - \hat c\|.
\]
The bound is vacuously zero when $\hat x = x^\star$.

 The event $\{\hat x \neq x^\star\}$ implies $\hat c$
and $c$ lie in different optimality cones, hence
$\|c-\hat c\|\geq \pi_v(c)$. Since $\pi_v$ is the distance to
a fixed closed union of hyperplanes, it is $1$-Lipschitz, so
\[
    \pi_v(c) = \pi_v(\theta^\star z + \varepsilon)
    \geq \pi_v(\theta^\star z) - \|\varepsilon\|
    \geq 1/\omega - \|\varepsilon\|.
\]
Combining with $\|c-\hat c\|\leq\|(\theta-\theta^\star)z\| +
\|\varepsilon\|$,
\[
    \{\hat x \neq x^\star\}
    \subseteq
    \bigl\{\|(\theta-\theta^\star)z\| + 2\|\varepsilon\|
    \geq 1/\omega\bigr\}.
\]

 Let $X := \|(\theta-\theta^\star)z\| + 2\|\varepsilon\|$.
Then by the above calculations
\[
    \mathrm{Regret} \leq L\,\|c-\hat c\|\,\mathbf{1}\{X\geq 1/\omega\}
    \leq L\,X\,\mathbf{1}\{X\geq 1/\omega\}
    \leq L\omega\,X^2,
\]
using $x\,\mathbf{1}\{x\geq 1/\omega\}\leq \omega x^2$ for $x\geq 0$.
By $(a+b)^2 \leq 2a^2 + 2b^2$,
$X^2 \leq 2\,\|(\theta-\theta^\star)z\|^2 + 8\|\varepsilon\|^2$.

Taking expectation and using
$\|(\theta-\theta^\star)z\|\leq R\,\|\theta-\theta^\star\|_F$ together
with $\mathbb{E}\|\varepsilon\|^2\leq d\,\sigma_\varepsilon^2$:
\[
    \mathbb{E}[\mathrm{Regret}]
    \leq L\omega\bigl(2\,\mathbb{E}\|(\theta-\theta^\star)z\|^2
    + 8\,\mathbb{E}\|\varepsilon\|^2\bigr)
    \leq 2L\omega R^2\,\mathbb{E}\|\theta-\theta^\star\|_F^{\,2}
    + 8L\omega\, d\,\sigma_\varepsilon^2. 
\] which give $C_1,C_2$.
\end{proof}

\subsubsection{Formal regret theorem statement}
\begin{theorem}[Validation-size regret regimes]\label{thm:regret-regimes}
Adopt Assumptions (D1)--(D5) and Assumptions ~\ref{ass:sb}-\ref{ass:bs}. 
Let $C_1 := 2L\omega R^2$ and $C_2 := 8L\omega d\sigma_\varepsilon^2$ be the constants
from Lemma~\ref{lem:regret-bridge}, and let $A_1^{\mathrm{unw}}, A_1^{\mathrm{rcif}},
A_1^{\mathrm{val}}, A_2^{\mathrm{unw}}, A_2^{\mathrm{rcif}}$ be the constants from
Lemma~\ref{lem:mse-rates}. Define the unconditional expected regret upper bounds
\begin{align*}
    R_{\mathrm{unw}}(n) &:= C_1\!\left(A_1^{\mathrm{unw}}\,\frac{\sigma_\varepsilon^2 d}{n\mu^2}
        + A_2^{\mathrm{unw}}\,\frac{\alpha^2\beta^2 R^2}{\mu^2}\right) + C_2,\\
    R_w(\lambda, m, n) &:= C_1\!\left(A_1^{\mathrm{rcif}}\,\frac{\sigma_\varepsilon^2 d}{n\mu^2}
        + A_2^{\mathrm{rcif}}\,\frac{\alpha^2\beta R^2\,\Phi(\lambda, m, n)}{\mu^2}\right) + C_2,\\
    R_{\mathrm{val}}(m) &:= C_1\!\left(A_1^{\mathrm{val}}\,\frac{\sigma_\varepsilon^2 d}{m\mu^2}\right) + C_2,
\end{align*}
where $\Phi(\lambda, m, n)$ is the
weight ratio bound in Lemma~\ref{lem:weight-ratio}. Define the thresholds
\[
    m_\star(\lambda) := \inf\!\left\{m \geq 1 \,:\, \Phi(\lambda, m, n) < \beta\,\frac{A_2^{\mathrm{unw}}}{A_2^{\mathrm{rcif}}}\right\},
\]
\[
    m^\star(\lambda) := \sup\!\left\{m \geq 1 \,:\, m \cdot \Phi(\lambda, m, n) < C\,\frac{A_2^{\mathrm{unw}}}{A_2^{\mathrm{rcif}}}\right\},
    \quad C := \frac{A_1^{\mathrm{unw}}\,\sigma_\varepsilon^2 d}{A_2^{\mathrm{unw}}\,\alpha^2\beta R^2}.
\]
Then in the asymptotic regime $n \to \infty$ with $m, \beta, \alpha, \lambda$ held fixed,
the following hold:

\noindent (1) (\emph{Lower bound on validation size: tighter expected regret bound for
RCIF vs.\ unweighted ERM.}) For all $m > m_\star(\lambda)$ and $n$ sufficiently large,
\[
    R_w(\lambda, m, n) < R_{\mathrm{unw}}(n) + o(1).
\]
The threshold $m_\star(\lambda)$ is finite whenever
$e^{-\lambda \alpha \gamma_0/4} < \beta\,A_2^{\mathrm{unw}}/A_2^{\mathrm{rcif}}$,
where $\gamma_0$ is the minimum-margin constant from Assumption~\ref{ass:sb}.
With the relationship $A_2^{\mathrm{rcif}} = e^{6\lambda M} A_2^{\mathrm{unw}}$ from
Lemma~\ref{lem:mse-rates}, this reads
$\lambda(\alpha \gamma_0/4 - 6M) > \ln(1/\beta)$, which holds for sufficiently large
$\lambda$ provided $\alpha \gamma_0 > 24M$.

\noindent (2) (\emph{Upper bound on validation size: tighter expected regret bound for
RCIF vs.\ validation-only ERM.}) For all $m < m^\star(\lambda)$ and $n$ sufficiently
large,
\[
    R_w(\lambda, m, n) < R_{\mathrm{val}}(m) + o(1).
\]
The threshold $m^\star(\lambda)$ is finite, since the bias term in $R_w$ stabilizes
to a positive constant as $m \to \infty$ while $R_{\mathrm{val}}(m) \to C_2$.

\noindent (3) (\emph{Sweet spot.}) If
\[
    \Phi(\lambda, m, n) < \beta\,\frac{A_2^{\mathrm{unw}}}{A_2^{\mathrm{rcif}}}
    \quad\text{and}\quad
    m \cdot \Phi(\lambda, m, n) < C\,\frac{A_2^{\mathrm{unw}}}{A_2^{\mathrm{rcif}}},
\]
then $m_\star(\lambda) < m^\star(\lambda)$, and for every $m \in (m_\star(\lambda), m^\star(\lambda))$
and $n$ sufficiently large,
\[
    R_w(\lambda, m, n) < \min\!\bigl(R_{\mathrm{unw}}(n),\, R_{\mathrm{val}}(m)\bigr) + o(1).
\]
\end{theorem}
\begin{proof}
By Lemma~\ref{lem:regret-bridge}, for any of the three estimators
$\hat\theta_\bullet \in \{\hat\theta, \hat\theta_p, \hat\theta_{\mathrm{val}}\}$,
\[
    \E[\mathrm{Regret}(\hat\theta_\bullet)] \leq C_1\,\E\|\hat\theta_\bullet - \theta^\star\|_F^2 + C_2.
\]
Combining with the method-specific MSE upper bounds of Lemma~\ref{lem:mse-rates} yields
the regret upper bounds $R_{\mathrm{unw}}, R_w, R_{\mathrm{val}}$ in the statement.

We use the following standing relationships among the constants, established in
Lemma~\ref{lem:mse-rates}:
\begin{itemize}
\item[(C1)] $A_1^{\mathrm{val}} = A_1^{\mathrm{unw}}$,
\item[(C2)] $A_1^{\mathrm{rcif}} = e^{6\lambda M}\,A_1^{\mathrm{unw}}$,
\item[(C3)] $A_2^{\mathrm{rcif}} = e^{6\lambda M} A_2^{\mathrm{unw}}$. 
\end{itemize}

\textbf{Part (1): comparison with $\hat\theta$.}
\[
    R_w(\lambda, m, n) - R_{\mathrm{unw}}(n)
    = \underbrace{C_1 (A_1^{\mathrm{rcif}} - A_1^{\mathrm{unw}})\,\frac{\sigma_\varepsilon^2 d}{n\mu^2}}_{=:\,(V)}
    + \underbrace{C_1\,\frac{\alpha^2 R^2}{\mu^2}\Bigl(A_2^{\mathrm{rcif}}\,\beta\,\Phi(\lambda, m, n) - A_2^{\mathrm{unw}}\,\beta^2\Bigr)}_{=:\,(B)}.
\]
The variance term $(V) = C_1(e^{6\lambda M} - 1)A_1^{\mathrm{unw}}\sigma_\varepsilon^2 d/(n\mu^2) = O(1/n)$,
hence absorbs into $o(1)$. The bias term $(B)$ is strictly negative iff
\[
    A_2^{\mathrm{rcif}}\,\beta\,\Phi(\lambda, m, n) < A_2^{\mathrm{unw}}\,\beta^2
    \;\Longleftrightarrow\;
    \Phi(\lambda, m, n) < \beta\,\frac{A_2^{\mathrm{unw}}}{A_2^{\mathrm{rcif}}}.
\]
By Lemma~\ref{lem:weight-ratio}, $\Phi(\lambda, m, n) \to e^{-\lambda \alpha \gamma_0/4} + o(1)$
as $m \to \infty$, so the set $\{m : \Phi(\lambda, m, n) < \beta A_2^{\mathrm{unw}}/A_2^{\mathrm{rcif}}\}$
is nonempty whenever $e^{-\lambda \alpha \gamma_0/4} < \beta A_2^{\mathrm{unw}}/A_2^{\mathrm{rcif}}$,
and $m_\star(\lambda)$ is well-defined. For all $m > m_\star(\lambda)$ and $n$ sufficiently
large, the comparison holds strictly, giving conclusion (1) up to $o(1)$.

\textbf{Part (2): comparison with $\hat\theta_{\mathrm{val}}$.} Using
$A_1^{\mathrm{val}} = A_1^{\mathrm{unw}}$ from (C1),
\[
    R_w(\lambda, m, n) - R_{\mathrm{val}}(m)
    = C_1 A_1^{\mathrm{rcif}}\,\frac{\sigma_\varepsilon^2 d}{n\mu^2}
    + C_1 A_2^{\mathrm{rcif}}\,\frac{\alpha^2\beta R^2\,\Phi(\lambda, m, n)}{\mu^2}
    - C_1 A_1^{\mathrm{unw}}\,\frac{\sigma_\varepsilon^2 d}{m\mu^2}.
\]
For $n$ large enough that $A_1^{\mathrm{rcif}}\sigma_\varepsilon^2 d/n$ is dominated by
the other two terms (absorbed into $o(1)$), the comparison reduces to
\[
    A_2^{\mathrm{rcif}}\,\frac{\alpha^2\beta R^2\,\Phi(\lambda, m, n)}{\mu^2}
    < A_1^{\mathrm{unw}}\,\frac{\sigma_\varepsilon^2 d}{m\mu^2}.
\]
Multiplying both sides by $m\mu^2/A_1^{\mathrm{unw}}$, this is equivalent to
\[
    m \cdot \Phi(\lambda, m, n) < C\,\frac{A_2^{\mathrm{unw}}}{A_2^{\mathrm{rcif}}},
    \quad C := \frac{A_1^{\mathrm{unw}}\,\sigma_\varepsilon^2 d}{A_2^{\mathrm{unw}}\,\alpha^2\beta R^2}.
\]
By definition this holds for all $m < m^\star(\lambda)$. To verify $m^\star(\lambda)$
is finite: by Lemma~\ref{lem:weight-ratio}, $\Phi$ stabilizes to the positive limit
$e^{-\lambda \alpha \gamma_0/4}$ as $m \to \infty$, so $f(m) := m \cdot \Phi(\lambda, m, n)$
grows at least linearly in $m$ for large $m$ and eventually exceeds the constant
$C\,A_2^{\mathrm{unw}}/A_2^{\mathrm{rcif}}$.

\textbf{Part (3): sweet spot.} Direct from Parts (1) and (2).
\end{proof}
\section{Additional experiments}
\label{app:exp}
\subsection{Corruption rate study shortest path}
\begin{figure}
    \centering
    \includegraphics[width=0.5\linewidth]{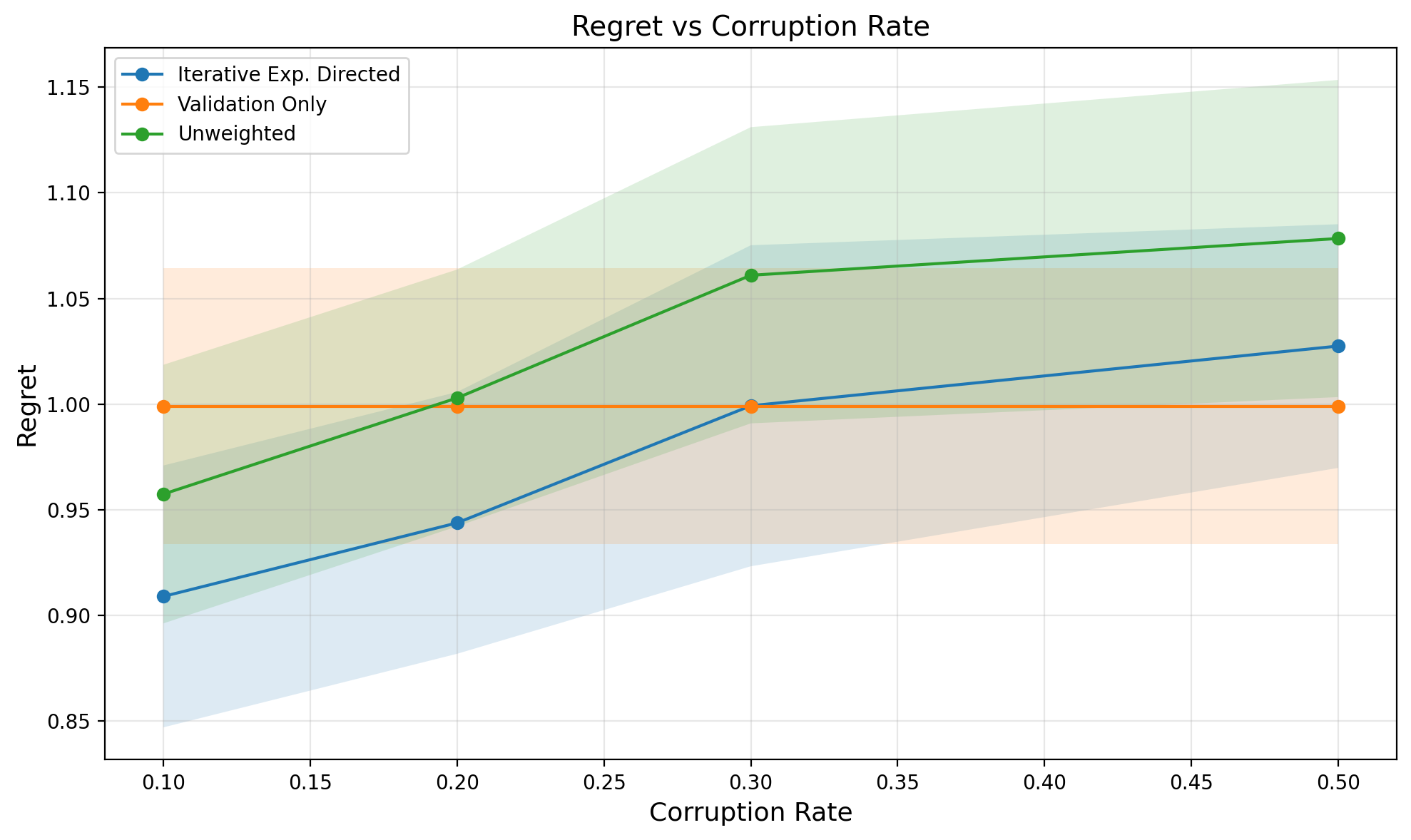}
    \caption{Corruption rate and regret}
    \label{fig:cor_rate}
\end{figure}

We vary the corruption rate $\beta$ across [0.1,0.5], observing that there is a similar tradeoff in corruption rate for $\hat \theta_p$ vs $\hat \theta_{\mathrm{val}}$ as seen in \Cref{fig:cor_rate}. As corruption rate increases, both unweighted and iterative exp. directed suffer in performance, while validation only remains constant. In particular, $\beta = 0.3$ appears to be the crossover point. This is to be expected, as once the dataset becomes sufficiently noisy, it is difficult to fit on, regardless of reweighting or not, but we do observe that our method consistently retains lower regret compared to the unweighted ERM. These experiments were run on shortest path problems with fixed parameters over 50 seeds: 4 $\times 4$ grid, upstream input dimension 30, 400 training points, 100 validation points, 200 test points.
The full table is given below in \Cref{tab:results_corruption}
\begin{table}[ht]
\centering
\small
\setlength{\tabcolsep}{4pt}
\begin{tabular}{lcccc}
\toprule
Method & $\beta=0.1$ & $\beta=0.2$ & $\beta=0.3$ & $\beta=0.5$ \\
\midrule
Iterative Exp. Directed & \textbf{0.9089} $\pm$ \textbf{0.0172} & \textbf{0.9437} $\pm$ \textbf{0.0172} & 0.9992 $\pm$ 0.0211 & 1.0275 $\pm$ 0.0160 \\
Exp. Two-Sided          & 0.9528 $\pm$ 0.0172 & 1.0009 $\pm$ 0.0175 & 1.0580 $\pm$ 0.0191 & 1.0817 $\pm$ 0.0208 \\
Directed                & 0.9357 $\pm$ 0.0170 & 0.9825 $\pm$ 0.0171 & 1.0433 $\pm$ 0.0189 & 1.0568 $\pm$ 0.0187 \\
Unweighted              & 0.9573 $\pm$ 0.0170 & 1.0029 $\pm$ 0.0169 & 1.0610 $\pm$ 0.0194 & 1.0784 $\pm$ 0.0208 \\
RC Weight               & 0.9575 $\pm$ 0.0171 & 1.0035 $\pm$ 0.0169 & 1.0604 $\pm$ 0.0192 & 1.0780 $\pm$ 0.0210 \\
Validation Only         & 0.9989 $\pm$ 0.0181 & 0.9989 $\pm$ 0.0181 & \textbf{0.9989} $\pm$ \textbf{0.0181} & \textbf{0.9989} $\pm$ \textbf{0.0181} \\
SPO Influence Weight & 0.9212 $\pm$ 0.0172 & 0.9560 $\pm$ 0.0158 & 0.9998 $\pm$ 0.0191 & 1.0307 $\pm$ 0.0209 \\
Clean ERM (oracle) & 0.8513 $\pm$ 0.0171 & 0.8513 $\pm$ 0.0171 & 0.8513 $\pm$ 0.0171 & 0.8513 $\pm$ 0.0171 \\
SPO+ Val Only & 1.0914 $\pm$ 0.0182 & 1.0914 $\pm$ 0.0182 & 1.0914 $\pm$ 0.0182 & 1.0914 $\pm$ 0.0182 \\
SPO+ Finetune & 1.0015 $\pm$ 0.0184 & 1.0042 $\pm$ 0.0176 & 1.0160 $\pm$ 0.0178 & 1.0233 $\pm$ 0.0190 \\
\bottomrule
\end{tabular}
\caption{Mean test regret as a function of corruption rate $\beta$. Best per column in bold.}
\label{tab:results_corruption}
\end{table}

\subsection{Validation study shortest path full details}
\label{app:ablvalfull}
We provide the performance of all methods across varying validation set sizes, not just the unweighted and validation only baselines. We see that RC Weight very slightly outperforms the unweighted baseline in performance particularly at smaller validation sizes, whereas our method excels starting in the medium validation set size range, with a larger margin. Results were obtained on the shortest path setting with 100 seeds on a $4 \times 4$ grid with upstream input dimension $20$, $500$ training points, $200$ test points, and $\beta = 0.2$, using the base setting from \Cref{tab:results_new_val}.
\begin{table}[ht]
\centering
\small
\setlength{\tabcolsep}{4pt}
\resizebox{\textwidth}{!}{%
\begin{tabular}{lccccccc}
\toprule
Method & |V| = 5 & |V| = 10 & |V| = 20 & |V| = 50 & |V| = 100 & |V| = 150 & |V| = 200 \\
\midrule
Iterative Exp. Directed & 1.0595 $\pm$ 0.0140 & 1.0051 $\pm$ 0.0127 & 0.9707 $\pm$ 0.0123 & 0.9354 $\pm$ 0.0120 & \textbf{0.9088 $\pm$ 0.0107} & \textbf{0.8751 $\pm$ 0.0107} & \textbf{0.8688 $\pm$ 0.0111} \\
Exp. Two-Sided & 1.0050 $\pm$ 0.0109 & 0.9935 $\pm$ 0.0116 & 0.9789 $\pm$ 0.0125 & 0.9586 $\pm$ 0.0135 & 0.9458 $\pm$ 0.0118 & 0.9228 $\pm$ 0.0117 & 0.9127 $\pm$ 0.0124 \\
Directed & 1.0015 $\pm$ 0.0129 & 0.9756 $\pm$ 0.0111 & 0.9659 $\pm$ 0.0115 & 0.9520 $\pm$ 0.0124 & 0.9389 $\pm$ 0.0117 & 0.9071 $\pm$ 0.0116 & 0.8964 $\pm$ 0.0121 \\
Unweighted & 0.9857 $\pm$ 0.0119 & 0.9736 $\pm$ 0.0113 & 0.9714 $\pm$ 0.0119 & 0.9627 $\pm$ 0.0125 & 0.9516 $\pm$ 0.0122 & 0.9282 $\pm$ 0.0121 & 0.9184 $\pm$ 0.0124 \\
RC Weight & \textbf{0.9855 $\pm$ 0.0120} & \textbf{0.9732 $\pm$ 0.0112} & 0.9713 $\pm$ 0.0117 & 0.9634 $\pm$ 0.0126 & 0.9515 $\pm$ 0.0122 & 0.9280 $\pm$ 0.0120 & 0.9179 $\pm$ 0.0124 \\
Validation Only & 1.1854 $\pm$ 0.0137 & 1.1710 $\pm$ 0.0145 & 1.2471 $\pm$ 0.0155 & 1.0366 $\pm$ 0.0142 & 0.9563 $\pm$ 0.0105 & 0.8940 $\pm$ 0.0115 & 0.8871 $\pm$ 0.0107 \\
SPO Influence Weight & 1.0009 $\pm$ 0.0111 & 0.9771 $\pm$ 0.0124 & \textbf{0.9506 $\pm$ 0.0113} & \textbf{0.9257 $\pm$ 0.0123} & 0.9185 $\pm$ 0.0108 & 0.8864 $\pm$ 0.0113 & 0.8813 $\pm$ 0.0115 \\
Clean ERM (oracle) & 0.8256 $\pm$ 0.0094 & 0.8265 $\pm$ 0.0088 & 0.8279 $\pm$ 0.0108 & 0.8255 $\pm$ 0.0104 & 0.8129 $\pm$ 0.0094 & 0.8068 $\pm$ 0.0107 & 0.8178 $\pm$ 0.0108 \\
SPO+ Val Only & 1.1854 $\pm$ 0.0137 & 1.1710 $\pm$ 0.0145 & 1.2471 $\pm$ 0.0155 & 1.1180 $\pm$ 0.0133 & 1.0386 $\pm$ 0.0133 & 0.9778 $\pm$ 0.0133 & 0.9567 $\pm$ 0.0117 \\
SPO+ Finetune & 0.9982 $\pm$ 0.0121 & 1.0064 $\pm$ 0.0124 & 1.0234 $\pm$ 0.0130 & 0.9997 $\pm$ 0.0119 & 0.9657 $\pm$ 0.0111 & 0.9356 $\pm$ 0.0116 & 0.9248 $\pm$ 0.0106 \\
\bottomrule
\end{tabular}
}
\caption{Mean test regret as a function of validation set size $|\mathcal{V}|$. Best per column in bold.}
\label{tab:results_val_size}
\end{table}
We also provide the same setting with $\beta=0.3$ for \Cref{fig:valsize}, where we demonstrate the crossing over point of Validation Only which is readily apparent at the higher corruption rate.
\begin{table}[ht]
\centering
\small
\setlength{\tabcolsep}{4pt}
\resizebox{\textwidth}{!}{%
\begin{tabular}{lccccccc}
\toprule
Method & |V| = 5 & |V| = 10 & |V| = 20 & |V| = 50 & |V| = 100 & |V| = 150 \\
\midrule
Iterative Exp. Directed & 1.0964 $\pm$ 0.0146 & 1.0600 $\pm$ 0.0132 & 1.0254 $\pm$ 0.0126 & \textbf{0.9745 $\pm$ 0.0117} & \textbf{0.9334 $\pm$ 0.0107} & 0.9120 $\pm$ 0.0121 \\
Unweighted & \textbf{1.0234 $\pm$ 0.0133} & \textbf{1.0159 $\pm$ 0.0113} & \textbf{1.0137 $\pm$ 0.0116} & 0.9995 $\pm$ 0.0122 & 0.9873 $\pm$ 0.0114 & 0.9719 $\pm$ 0.0120 \\
Validation Only & 1.1854 $\pm$ 0.0137 & 1.1710 $\pm$ 0.0145 & 1.2471 $\pm$ 0.0155 & 1.0366 $\pm$ 0.0142 & 0.9563 $\pm$ 0.0105 & \textbf{0.8940 $\pm$ 0.0115} \\
\bottomrule
\end{tabular}
}
    \caption{Mean test regret values across $|\mathcal{V}|$ for Fig.~1}
    \label{tab:placeholder}
\end{table}
\subsection{Corruption rate study for random LP}
We provide a similar study on corruption rate, now with the random LP downstream. The gaps between the reweighting methods are small: Iterative Exp.\ Directed and SPO Influence Weight (paired $p \geq 0.18$), and both improve on unweighted ERM by a modest margin that is significant at $\beta = 0.3$ ($p = 0.016$). Validation Only is constant across $\beta$ and attains the lowest mean regret at $\beta = 0.5$ as much of the data has been corrupted, but the difference to our method is within noise ($p = 0.42$); the clean-ERM ceiling shows that the total damage from corruption is small in this setting ($0.025$ to $0.08$ regret), which bounds what any reweighting can recover. The SPO+ baselines do not help here: fine-tuning on the validation set is worse than unweighted ERM at every $\beta$, and training on the validation set alone is far worse. Results were obtained with upstream input dimension $10$, LP dimension $10$, $20$ constraints, train set size $200$, validation set size $50$, test set size $100$, and $100$ seeds.
\begin{table}[ht]
\centering
\small
\setlength{\tabcolsep}{4pt}
\resizebox{\textwidth}{!}{%
\begin{tabular}{lcccc}
\toprule
Method & $\beta$ = 0.1 & $\beta$ = 0.2 & $\beta$ = 0.3 & $\beta$ = 0.5 \\
\midrule
Iterative Exp. Directed & 0.3746 $\pm$ 0.0176 & 0.3865 $\pm$ 0.0180 & 0.3978 $\pm$ 0.0195 & 0.4230 $\pm$ 0.0181 \\
Exp. Two-Sided & 0.3774 $\pm$ 0.0181 & 0.3893 $\pm$ 0.0182 & 0.4037 $\pm$ 0.0185 & 0.4331 $\pm$ 0.0180 \\
Directed & 0.3771 $\pm$ 0.0182 & 0.3896 $\pm$ 0.0180 & 0.4010 $\pm$ 0.0187 & 0.4341 $\pm$ 0.0184 \\
Unweighted & 0.3764 $\pm$ 0.0181 & 0.3896 $\pm$ 0.0186 & 0.4053 $\pm$ 0.0183 & 0.4333 $\pm$ 0.0179 \\
RC Weight & 0.3772 $\pm$ 0.0186 & 0.3906 $\pm$ 0.0190 & 0.4046 $\pm$ 0.0184 & 0.4333 $\pm$ 0.0178 \\
Validation Only & 0.4173 $\pm$ 0.0197 & 0.4173 $\pm$ 0.0197 & 0.4173 $\pm$ 0.0197 & \textbf{0.4173} $\pm$ \textbf{0.0197} \\
SPO Influence Weight & \textbf{0.3709} $\pm$ \textbf{0.0176} & \textbf{0.3863} $\pm$ \textbf{0.0181} & \textbf{0.3957} $\pm$ \textbf{0.0193} & 0.4251 $\pm$ 0.0192 \\
Clean ERM (oracle) & 0.3513 $\pm$ 0.0157 & 0.3513 $\pm$ 0.0157 & 0.3513 $\pm$ 0.0157 & 0.3513 $\pm$ 0.0157 \\
SPO+ Val Only & 0.6970 $\pm$ 0.0337 & 0.6970 $\pm$ 0.0337 & 0.6970 $\pm$ 0.0337 & 0.6970 $\pm$ 0.0337 \\
SPO+ Finetune & 0.4357 $\pm$ 0.0184 & 0.4455 $\pm$ 0.0201 & 0.4407 $\pm$ 0.0188 & 0.4694 $\pm$ 0.0210 \\
\bottomrule
\end{tabular}
}
\caption{Mean test regret as a function of corruption rate $\beta$ on the random LP setting. Best per column in bold.}
\label{tab:results_corruption_randomlp}
\end{table}
\subsection{Validation size study for random LP}
We provide a similar study on validation set size, but for random LP downstream. Like we have previously observed, the impact of RCIF reweighting is less apparent in this setting (although still present). In particular, it appears that the range of validation set sizes for which the RCIF scoring performs better than unweighted is smaller, which is likely due to the lack of structure in the downstream problem. Results were obtained with upstream input dimension 10, LP dimension 10, 20 constraints, train set size 200, test set size 100, and corruption rate $\beta = 0.5$.
\begin{table}[ht]
\centering
\small
\setlength{\tabcolsep}{4pt}
\resizebox{\textwidth}{!}{%
\begin{tabular}{lcccccc}
\toprule
Method & |V| = 5 & |V| = 10 & |V| = 20 & |V| = 50 & |V| = 100 & |V| = 150 \\
\midrule
Iterative Exp. Directed & 0.6033 $\pm$ 0.0331 & 0.5311 $\pm$ 0.0258 & 0.4640 $\pm$ 0.0221 & 0.4230 $\pm$ 0.0181 & 0.3693 $\pm$ 0.0185 & 0.3632 $\pm$ 0.0160 \\
Exp. Two-Sided & 0.4774 $\pm$ 0.0255 & 0.4704 $\pm$ 0.0225 & \textbf{0.4491 $\pm$ 0.0209} & 0.4331 $\pm$ 0.0180 & 0.3848 $\pm$ 0.0183 & 0.3758 $\pm$ 0.0164 \\
Directed & 0.4766 $\pm$ 0.0251 & 0.4667 $\pm$ 0.0227 & 0.4498 $\pm$ 0.0207 & 0.4341 $\pm$ 0.0184 & 0.3835 $\pm$ 0.0183 & 0.3737 $\pm$ 0.0165 \\
Unweighted & \textbf{0.4576 $\pm$ 0.0232} & 0.4618 $\pm$ 0.0219 & 0.4503 $\pm$ 0.0203 & 0.4333 $\pm$ 0.0179 & 0.3830 $\pm$ 0.0183 & 0.3752 $\pm$ 0.0165 \\
RC Weight & 0.4596 $\pm$ 0.0236 & \textbf{0.4609 $\pm$ 0.0218} & 0.4501 $\pm$ 0.0208 & 0.4333 $\pm$ 0.0178 & 0.3838 $\pm$ 0.0184 & 0.3750 $\pm$ 0.0168 \\
Validation Only & 9.8179 $\pm$ 0.4086 & 7.9994 $\pm$ 1.3074 & 0.6616 $\pm$ 0.0284 & \textbf{0.4173 $\pm$ 0.0197} & \textbf{0.3570 $\pm$ 0.0190} & \textbf{0.3602 $\pm$ 0.0159} \\
SPO Influence Weight & 0.4712 $\pm$ 0.0243 & 0.4722 $\pm$ 0.0223 & 0.4593 $\pm$ 0.0204 & 0.4251 $\pm$ 0.0192 & 0.3757 $\pm$ 0.0185 & 0.3690 $\pm$ 0.0161 \\
Clean ERM (oracle) & 0.3447 $\pm$ 0.0165 & 0.3511 $\pm$ 0.0152 & 0.3451 $\pm$ 0.0163 & 0.3513 $\pm$ 0.0157 & 0.3357 $\pm$ 0.0175 & 0.3437 $\pm$ 0.0151 \\
SPO+ Val Only & 9.8179 $\pm$ 0.4086 & 7.9994 $\pm$ 1.3074 & 0.7749 $\pm$ 0.0360 & 0.6970 $\pm$ 0.0337 & 0.6588 $\pm$ 0.0296 & 0.5773 $\pm$ 0.0266 \\
SPO+ Finetune & 0.4747 $\pm$ 0.0233 & 0.4775 $\pm$ 0.0219 & 0.4878 $\pm$ 0.0219 & 0.4694 $\pm$ 0.0210 & 0.4141 $\pm$ 0.0188 & 0.4053 $\pm$ 0.0185 \\
\bottomrule
\end{tabular}
}
\caption{Mean test regret as a function of validation set size $|\mathcal{V}|$ on the random LP setting. Best per column in bold.}
\label{tab:results_val_size_randomlp}
\end{table}
\subsection{Adversarial noise}
\label{app:advers}
Beyond the isotropic Gaussian corruption considered in the main
experiments, we also evaluate our method against a more targeted
adversary that uses minimum-norm perturbations to flip the optimality
cone of corrupted training points. This corruption model is
particularly challenging for influence-based methods because it is
designed to induce wrong downstream decisions while remaining as
close as possible to the clean label.

\paragraph{Construction.} For a corrupted index $j \in \mathcal{C}$
with clean cost $c_j$, let $B_j$ be the optimal basis for $c_j$ and
let $N(B_j)$ denote the corresponding nonbasic set. The minimum
$\ell_2$-norm perturbation that pushes $c_j$ across the boundary of
$\Lambda(B_j)$ along nonbasic direction $k$ has magnitude
$s_k(c_j, B_j) / \|\delta(B_j, k)\|$. The cheapest direction to flip is
\[
    k^\dagger_j
    \;=\;
    \arg\min_{k \in N(B_j)}\,
    \frac{s_k(c_j, B_j)}{\|\delta(B_j, k)\|}.
\]
Given a budget $\rho > 0$, the adversarial corruption is
\[
    \tilde c_j
    \;=\;
    c_j - \alpha_j\,
    \frac{\delta(B_j, k^\dagger_j)}{\|\delta(B_j, k^\dagger_j)\|},
    \qquad
    \alpha_j = \min\!\left(\rho,\;
        \frac{(1+\varepsilon)\, s_{k^\dagger_j}(c_j, B_j)}
             {\|\delta(B_j, k^\dagger_j)\|}\right),
\]
with a small slack $\varepsilon > 0$ to ensure the perturbation
strictly crosses the boundary when budget allows, which is similar to the directional attack in Assumption D4, but with a variable magnitude. If $\rho$ is
insufficient to flip any direction, we use the full budget along the
cheapest flip direction without crossing.

Note that the perturbation magnitude is \emph{minimal}
rather than fixed: the adversary uses just enough perturbation to
cross a cone boundary, producing small residuals at the clean ERM
estimator and weak first-order influence signals.

\paragraph{Empirical observations.} Under cone-flipping corruption,
all methods including ours perform comparably to unweighted ERM, with
no method showing a significant improvement over unweighted
(Table~\ref{tab:adv}), perhaps due to the magnitude of the corruption being small so they perform similarly to unweighted. Diagnostic analysis confirms that the
RCIF score does carry information distinguishing corrupted from clean
points: the mean score over corrupted indices is consistently more
negative than over clean indices but the separation is insufficient for reliable detection. We
attribute this to a fundamental property of first-order influence
methods: the influence of a training point scales with its
perturbation magnitude, and minimum-norm adversarial corruption is
designed precisely to minimize this signal. Methods based on residual
magnitude or influence function approximations are therefore
inherently limited against such adversaries, and we expect that
defenses against cone-flipping corruption require either
second-order influence information or outlier-detection such as \cite{diakonikolas2019sever}. Results were obtained with both random LP and shortest path problems. For random LP, settings were upstream input dimension 10, LP dimension 10, 20 constraints, train set size 200, test set size 100, validation set size 50, $\beta = 0.3$, $\rho = 0.3$. For shortest path settings: grid 5 $\times$ 5, upstream input dimension 30, train set size 300, test set size 200, validation set size 100, $\beta = 0.3$, $\rho = 0.05$. $\rho$ is much smaller for shortest path because the optimality cones appear to be narrower, and thus require less perturbation to flip.
\begin{table}[ht]
\centering
\small
\begin{tabular}{lcc}
\toprule
Method & Random LP & Shortest Path \\
\midrule
Iterative Exp. Directed & 0.3527 $\pm$ 0.0158 & 1.2412 $\pm$ 0.0204 \\
Exp. Two-Sided & \textbf{0.3514 $\pm$ 0.0155} & 1.2430 $\pm$ 0.0196 \\
Directed & 0.3514 $\pm$ 0.0155 & 1.2424 $\pm$ 0.0195 \\
Unweighted & 0.3522 $\pm$ 0.0156 & 1.2412 $\pm$ 0.0194 \\
RC Weight & 0.3519 $\pm$ 0.0155 & \textbf{1.2410 $\pm$ 0.0194} \\
Validation Only & 0.4173 $\pm$ 0.0197 & 1.4459 $\pm$ 0.0177 \\
SPO Influence Weight & 0.3524 $\pm$ 0.0157 & 1.2410 $\pm$ 0.0199 \\
Clean ERM (oracle) & 0.3513 $\pm$ 0.0157 & 1.2422 $\pm$ 0.0196 \\
SPO+ Val Only & 0.6970 $\pm$ 0.0337 & 1.5713 $\pm$ 0.0207 \\
SPO+ Finetune & 0.4264 $\pm$ 0.0179 & 1.3983 $\pm$ 0.0203 \\
\bottomrule
\end{tabular}
\caption{Mean test regret under adversarial corruption (minimal magnitude). Best per column in bold.}
\label{tab:adv}
\end{table}
\subsubsection{Adversarial noise with fixed magnitude and robust regression}
\label{app:robust}

Residual-based robust regression is the natural alternative to reweighting when training labels are corrupted, so we compare against two standard instances: Huber regression (threshold $1.35$, fit per output coordinate) and iteratively trimmed least squares (three rounds, each refitting after discarding the training points with the largest residual norms). Both are fit on the same train-plus-validation pool as unweighted ERM. For trimmed ERM we set the trim fraction equal to the true corruption rate $\beta$, which is an oracle choice that favors the baseline. We evaluate under two corruption models of the same magnitude scale: the isotropic Gaussian corruption of the main experiments and the directional corruption of Assumption~(D4), in which each corrupted label is shifted by a fixed amount $\alpha$ along the most fragile direction of its own optimality cone. This is the same adversarial noise direction as \Cref{app:advers} but with a \textit{fixed} magnitude, not minimal. Both are run on the shortest-path problem ($4 \times 4$ grid, input dimension $20$, $400/100/200$ train/validation/test points, $50$ seeds), with $\beta = 0.2$ for the Gaussian setting and $\beta = 0.4$, $\alpha = 1.5$ for the directional setting. Table~\ref{tab:robust} reports the results under various types of noise.

Under isotropic Gaussian corruption the robust methods are strong, and trimmed ERM is the best method in every column: it attains $0.8544 \pm 0.0141$ against $0.9085 \pm 0.0164$ for Iterative Exp.\ Directed and $0.9600 \pm 0.0158$ for unweighted ERM, and it also improves on our method similarly for random LP (paired Wilcoxon $p < 0.001$ in each case) while Huber loss is consistent but shows the greatest gains under random LPs. We note that this is the regime these methods are designed for, and furthermore these methods have been given the corruption rate, an advantage our method nor any other method receives. Gaussian corruption of scale $2$ produces residuals that are large relative to the label noise, so a corrupted point is visible in residual norm at the unweighted fit, and trimming at the true rate removes most of them; Huber regression down-weights the same points more gently and lands between unweighted ERM and our method on the shortest-path settings ($0.9266 \pm 0.0163$).

Under directional corruption the ordering reverses. The shift $\alpha = 1.5$ is small relative to the label noise and does not stand out in residual norm, but it is aimed at the fragile direction of the cone and flips the optimal decision of every corrupted point. Trimmed ERM now discards clean points as often as corrupted ones and reaches $0.8977 \pm 0.0158$, worse than unweighted ERM ($0.8591 \pm 0.0164$) on $84\%$ of seeds ($p < 0.001$); Huber reaches $0.8730 \pm 0.0156$, also worse than unweighted ERM ($p < 0.001$). Iterative Exp.\ Directed is better at $0.8509 \pm 0.0164$, improving on unweighted ERM ($p = 0.036$) and on both robust baselines ($p < 0.001$ for each). The total damage is small in this setting (clean ERM attains $0.8257 \pm 0.0152$), which compresses all gaps, but the sign of the effect is what matters: the robust losses lose ground precisely when the corruption is decision-targeted rather than magnitude-visible.

The two experiments isolate different notions of robustness. Residual-based methods are robust to corruption magnitude and require that corrupted labels be outliers in the regression sense, together with a well-specified model and, for trimming, knowledge of the corruption rate and noise type. RCIF reweighting is robust to decision-targeted corruption: it scores a training point by its first-order effect on the reduced-cost margins at the clean validation points, which responds to where a label pushes the downstream decision rather than to how far it moves the label, and it requires neither the corruption rate nor magnitude-visible residuals. We regard the two approaches as complementary, and combining RCIF scores with a robust training loss is a natural direction for future work.
\begin{table}
    \centering
    \begin{tabular}{lccc}
        \toprule
        Method & LP, Gauss. $\beta$ = 0.3 & Shortest, Gauss. $\beta$ = 0.2 & Shortest, Directional $\beta$ = 0.4 \\
        \midrule
        Iterative Exp. Directed & 0.3451 $\pm$ 0.0133 & 0.9085 $\pm$ 0.0164 & \textbf{0.8509} $\pm$ 0.0164 \\
        Directed & 0.3531 $\pm$ 0.0129 & 0.9459 $\pm$ 0.0168 & 0.8547 $\pm$ 0.0161 \\
        Unweighted & 0.3564 $\pm$ 0.0130 & 0.9600 $\pm$ 0.0158 & 0.8591 $\pm$ 0.0164 \\
        Clean ERM (oracle) & 0.2785 $\pm$ 0.0109 & 0.8257 $\pm$ 0.0152 & 0.8257 $\pm$ 0.0152 \\
        Huber & 0.2952 $\pm$ 0.0108 & 0.9266 $\pm$ 0.0163 & 0.8730 $\pm$ 0.0156 \\
        Trimmed ERM & \textbf{0.2865} $\pm$ 0.0118 & \textbf{0.8544} $\pm$ 0.0141 & 0.8977 $\pm$ 0.0158 \\
        \bottomrule                               
    \end{tabular}
    
    \caption{Comparison against robust regression methods; under standard Gaussian isotropic noise they outperform every method, but under directional shifts greatly underperform the unweighted approach, requiring specific noise structure.}
    \label{tab:robust}
\end{table}

\subsection{Degeneracy}
Note that shortest-path and combinatorial problems generally violate primal nondegeneracy. For LPs with generic continuous $b$ and full-row-rank $A$ (such as our random-LP setting, or pricing/portfolio-style problems with continuous coefficients), degeneracy is a measure-zero event, so the assumption holds almost surely. More importantly, the assumption serves the theory rather than the algorithm: the method still functions and applies to degenerate problems, at the cost of some conservatism.

The algorithm itself never uses nondegeneracy: it solves one LP per validation point under the \emph{predicted} cost, takes whichever optimal basis the solver returns, and determines correctness via the cone test $s_k(c_v, B_v) \geq 0\ \ \forall k \in N(B_v)$, which is well defined under degeneracy. The cost of degeneracy is conservatism: the containment $\Lambda(B_v) \subseteq \mathcal{N}(x^*(\hat{c}_v))$ becomes strict, so the algorithm may mark a correct decision as incorrect when $c_v$ certifies the same vertex through a different basis, and the computed margin is a lower bound on the true decision margin. We note that (i) the correctness certificate never certifies a truly suboptimal point; and (ii) for the directed variant, such a misclassification is harmless at the point in question, because the violated directions there are all degenerate pivots. Its cost is signal efficiency, since part of the validation signal is spent
on basis bookkeeping rather than on actual decisions. Empirically, our gains are largest on shortest path, which we take as evidence that the conservative signal remains strong where the assumption fails

\subsection{Experimental details}
\subsubsection{Hardware and software}
All experiments were conducted on a personal laptop equipped with an Intel i7 CPU, 16GB RAM, and no dedicated GPU. Implementation required standard Python packages such as PyTorch, scikit-learn, as well as Gurobipy and Gurobi software for the optimization aspects. Each seed (which involves all baselines and optimization calls) runs in approximately 4 minutes. Code for reproducing our experiments is available at: https://github.com/william-zhang64/Reduced-Cost-Influence-Functions.
\subsubsection{Parameter and baseline settings}
Throughout the experiments, we kept $\lambda$ fixed at 0.05. The values $\lambda = 0.05$ and $\tau = 5$ were selected during preliminary experimentation across a grid of values $\{0.01,0.05,0.1,0.5\}$ and $\{3,5,10\}$ respectively, with a plateau in performance for $\lambda\in\{0.05,0.1\}$. Qualitatively, larger $\lambda$ produced overly extreme weights and degraded test regret, consistent with overfitting the small clean validation set, while smaller $\lambda$ left the weights too close to uniform to counteract the corruption. The number of iterations behaves similarly: beyond roughly $\tau = 5$ we observed a plateau in performance. Given that the pair $(\lambda, \tau) = (0.05, 5)$ was used across the settings in the paper for strong performance, this suggests these values serve as reasonable defaults.  Note that the weights were normalized such that unweighted had each weight = 1.0, so the total weights sum to $n$ rather than 1, which does not substantively affect any properties of the algorithm. Similarly, the $\phi$ parameter for the Directed method was also fixed to 0.05, for similar reasons. We used $\tau = 5$ iterations across the experiments for the Iterative Exp. Directed method. For SPO influence, we set the number of iterations to 5 as well, with learning rate of 0.01. For SPO+ Val Only we run full-batch subgradient descent on the SPO+ loss over the validation set for 100 epochs with normalized subgradients and diminishing step size $0.5/\sqrt{t+1}$, initialized at the validation-only least-squares fit, and keep the iterate with the lowest SPO+ objective. SPO+ Finetune uses the same procedure initialized at the unweighted least-squares fit on the train-plus-validation pool, with 30 epochs and step size $0.1/\sqrt{t+1}$. Huber regression uses threshold 1.35 and is fit per output coordinate on the train-plus-validation pool, and trimmed ERM runs three rounds, each refitting after discarding the fraction $\beta$ of training points with the largest residual norms. For the random LP data-generating process, we used standard normal distribution for features. For the isotropic noise, we used a covariance matrix of 4$\mathcal{I}_d$ and for the directional noise with fixed magnitude we used $\alpha=1.5$. For the neural network, we used hidden layer of size 8, with learning rate 1e-3.

\subsubsection{Computational cost}
\label{app:complexity}

Each iteration of RCIF reweighting solves the downstream LP once at each of the $m$ validation points, computes the influence functions of the $n$ training points, and refits the predictor by weighted least squares. The LP is never differentiated through and is not solved at training points. For the linear predictor with squared loss the influence has the closed form of Lemma~\ref{lem:factorization}: the Hessian is $2 I_d \otimes \hat\Sigma$, so one $q \times q$ factorization is shared by all training and validation points. Forming $\hat\Sigma$, its inverse, and the residuals $r_j = \hat\theta z_j - \tilde c_j$ costs $O(nq(q+d) + q^3)$ per iteration. At a validation point $v$ with direction $\delta_k$, the scores of all $n$ training points are the elementwise product of $\delta_k^\top r_j$, an $n \times d$ matrix-vector product, and $\kappa(z_v, z_j) = z_j^\top \hat\Sigma^{-1} z_v$, an $n \times q$ one, so scoring over the validation set costs $O(mn(q+d))$; the weighted least-squares refit on the $n + m$ points is absorbed into the same terms. The LP solves and basis information cost $O(m\,T_{\mathrm{LP}})$ per iteration, where $T_{\mathrm{LP}}$ is the cost of one simplex solve, and the iterative variant repeats all of this $\tau$ times. In practice the $\tau m$ LP solves dominate: on a laptop CPU with Gurobi, in the shortest-path setting of Table~\ref{tab:results_new_val} ($n = 400$, $m = 100$), one scoring round takes about $0.5$\,s and the complete Iterative Exp.\ Directed fit with $\tau = 5$ about $2$\,s.

\subsubsection{SPO Influence}
\label{app:spo-baseline}
We adapt the influence-based reweighting framework of
\citet{ren2018learning} to the predict-then-optimize setting by
substituting the SPO+ surrogate loss \citep{elmachtoub2022smart}
for validation MSE. This isolates the contribution of LP cone
geometry in our scoring from the general benefit of influence-based
reweighting on a downstream-aware loss.

For predicted cost $\hat{c}_v = f_{\hat\theta}(z_v)$ and true cost
$c_v$, SPO+ is a convex upper bound on regret with subgradient
\[
    g(\hat{c}_v, c_v)
    \;=\;
    2\big(x^*(c_v) - x^*(2\hat{c}_v - c_v) \big),
\]
where $x^*(c) = \arg\min_{x \in \mathcal{X}} c^\top x$.
Composing with the influence function $\mathcal{I}_{\hat\theta}(w_j)$
gives the per-training-point sensitivity of validation loss:
\[
    \Sigma^{\text{SPO+}}_j
    \;=\;
    \frac{1}{m}
    \sum_{w_v \in \mathcal{V}}
    g(\hat{c}_v, c_v)^\top \,
    \nabla_\theta f_{\hat\theta}(z_v)\,
    \mathcal{I}_{\hat\theta}(w_j).
\]
Starting from uniform weights $p^{(0)}_j = 1$, we iteratively update
\[
    p^{(t+1)}_j
    \;\leftarrow\;
    \max\!\big(0,\; p^{(t)}_j - \xi\,\Sigma^{\text{SPO+}}_j\big|_{\hat\theta^{(t)}}\big),
\]
normalize so that $\sum_j p^{(t+1)}_j = n$, and refit $\hat\theta^{(t+1)}$
via weighted ERM. We use $\xi = 0.01$ and run for $\tau = 5$ iterations.

This baseline matches RCIF in data usage and influence machinery; the
methods differ only in the projection direction for the prediction
shift: $g(\hat{c}_v, c_v)$ for SPO+ versus the simplex direction
$\delta(B_v, k)$ for RCIF.


\end{document}